\documentclass[10pt]{article} % For LaTeX2e
\usepackage[preprint]{tmlr}
\usepackage{hyperref}
\usepackage{url}

\usepackage{hyperref}
\usepackage{url}
\usepackage{graphicx} 
\usepackage{booktabs}
\usepackage{colortbl}
\usepackage{amssymb}
\usepackage{bbding}
\usepackage{pifont}
\usepackage{amsthm}
\usepackage{chngcntr}
\counterwithout{table}{section}
\usepackage{amsmath} 
\newtheorem{lemma}{Lemma}          % 独立编号的 lemma
\newtheorem{proposition}{Proposition} % 独立编号的 proposition
\theoremstyle{definition}

\usepackage[utf8]{inputenc}
\usepackage{algorithm}
\usepackage{algpseudocode}
\usepackage{amsmath}
\usepackage{caption}

\title{Distributionally Robust Alignment for Medical Federated Vision-Language Pre-training Under Data Heterogeneity}

\author{Zitao Shuai, Chenwei Wu, Zhengxu Tang \& Liyue Shen \thanks{coresponding author} \\
University of Michigan\\
\texttt{\{liyues\}@umich.edu}
}

\def\month{MM}  % Insert correct month for camera-ready version
\def\year{YYYY} % Insert correct year for camera-ready version
\def\openreview{\url{https://openreview.net/forum?id=XXXX}} % Insert correct link to OpenReview for camera-ready version

\begin{document}

\maketitle

\begin{abstract}
Vision-language pre-training (VLP) has emerged as an effective scheme for multimodal representation learning, but its reliance on large-scale multimodal data poses significant challenges for medical applications. Federated learning (FL) offers a promising solution to scale up the dataset for medical VLP while preserving data privacy. However, we observe that client data heterogeneity in real-world scenarios could cause models to learn biased cross-modal alignment during local pre-training. This would limit the transferability of the federally learned representation model on downstream tasks. To address this challenge, we propose \textbf{Fed}erated \textbf{D}istributionally \textbf{R}obust \textbf{A}lignment (FedDRA), a framework for federated VLP that achieves robust vision-language alignment under heterogeneous conditions.
Based on client datasets, we construct a distribution family that encompasses potential test-time domains, and apply a distributionally robust framework to optimize the pre-trained model's performance across this distribution space. This approach bridges the gap between pre-training samples and downstream applications. To avoid over-fitting on client-specific information, we use anchor representation from the global model to guide the local training, and adopt a two-stage approach to first tune deeper layers before updating the entire network. Extensive experiments on real-world datasets demonstrate FedDRA’s effectiveness in enhancing medical federated VLP under data heterogeneity. Our method also adapts well to various medical pre-training methods.
\end{abstract}

 %Specifically, to bridge the gap between downstream testing samples and pre-training datasets, we construct a distribution family based on client datasets, and employ a decentralized distributionally robust optimization method, to iteratively improve pre-trained model's performance on it. Moreover, to alleviate the negative affect of client-specific information on learning cross-modal alignment, we maintain a global model to provide anchor representations for guiding local training, and utilize a two-stage training schema to separately update deep layers and shallow layers. 

\section{Introduction}
\label{introduction}

Vision-language pre-training (VLP) learns transferable multimodal representations by extracting latent semantics from large-scale image-text pairs, where the dataset scale largely impacts the performance of the learned model~\citep{oquab2023dinov2}. 
However, scaling up multimodal pre-training datasets is a non-trivial challenge especially for medical applications, due to privacy concerns and regulations of patient data sharing~\citep{ladbury2023integration}. Recent work has explored federated learning as a solution to leverage data across multiple medical institutions while preserving privacy~\citep{lu2023scaling}.

\begin{figure}[!t]
    \centering
    % \vspace{-100pt}
    \includegraphics[width=\textwidth]{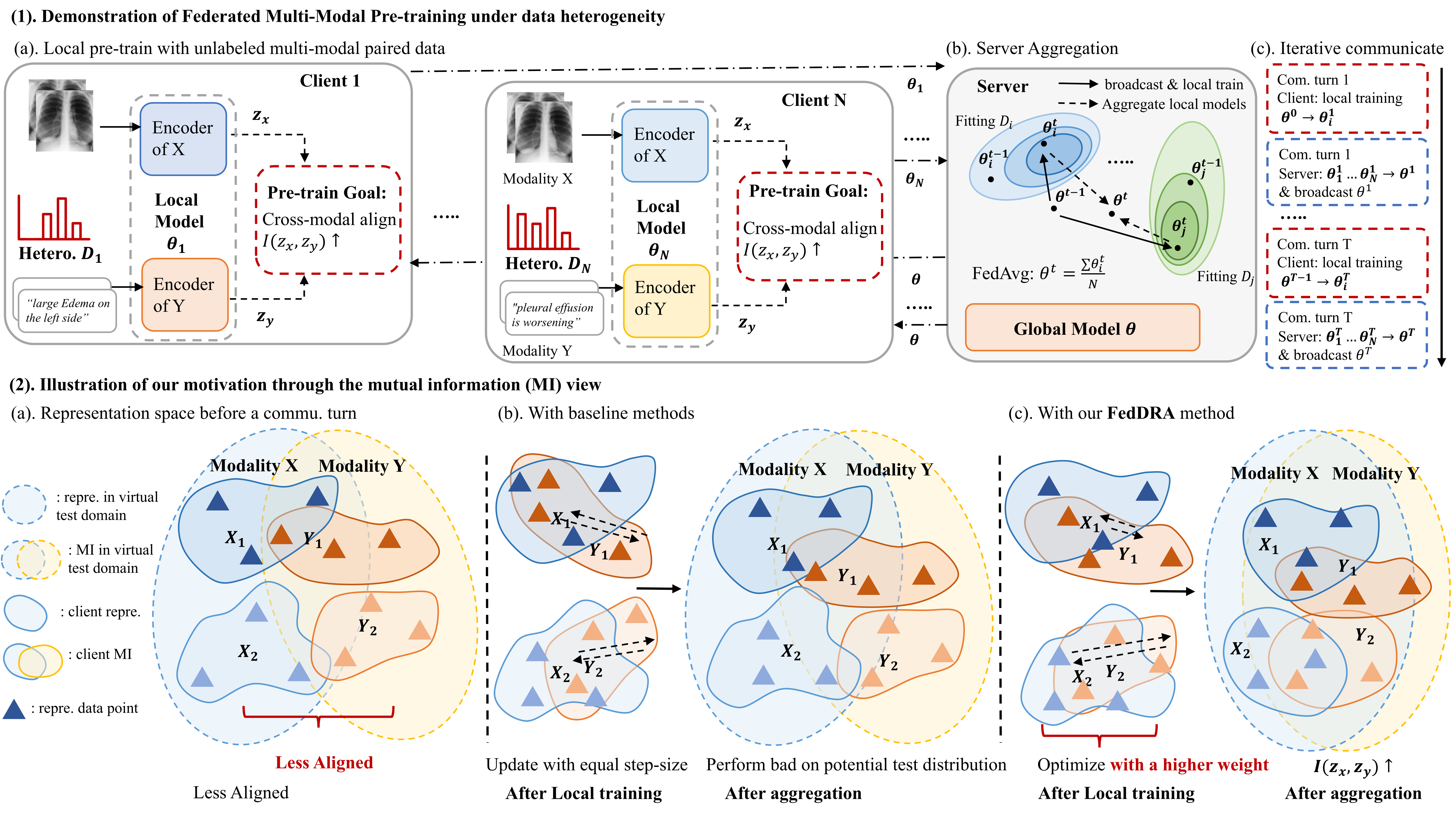} % Adjust the path and width as needed
    \vspace{-10pt}
    \caption{(1) We aim to tackle the data-hungry problem in VLP via federally utilizing private multi-modal paired data. Pre-training local models on heterogeneous client datasets may overfit the observed local data.  (2). The naive method ignores the disparity between local distribution and potential testing distribution, and directly averaging local pre-trained models would obtain a less generalizable model. In contrast, our method optimizes model performance based \textit{on a family of potential testing distributions}, dynamically given weights to schedule the local training, and obtain a more transferable model.
    }
    \label{fig:intro}
    \vspace{-10pt}
\end{figure}

However, in real-world scenarios, datasets collected from different institutes are always heterogeneous. For example, hospitals in tropical regions receive a high proportion of pneumonia patients, whereas those in colder climates may see more pneumothorax cases
~\citep{mendogni2020epidemiology}. This data heterogeneity is not only a long-standing problem in classical federated learning~\citep{ghosh2019robust,huang2022learn}, but a practical challenge that impedes the deployment of medical VLP in the federated learning setting. Current medical VLP methods often focus on learning a modality-shared latent space, where their training multi-modal data pairs are well-aligned. However, such learned cross-modal alignments may not be transferable to data from unseen distributions. As shown in Fig.~\ref{fig:intro}, this will harm the performance of the federally pre-trained model, which is aggregated from client models trained on heterogeneous local datasets.

We start by investigating how data heterogeneity affects the performance of federally pre-trained VL models. In classic medical VLP~\citep{wang2022multi, bannur2023learning}, the model learns cross-modal alignment through maximizing the mutual information of the two modalities on its observed training data. In federated settings, as shown in Fig.~\ref{fig:intro}, this approach often learns local models that overfit client-specific information. Also, averaging these local models' parameters cannot always produce a model with generalizable cross-modal alignment. Secondly, biased deep layers, which overfit multi-modal correlations of local datasets, would prevent the model from learning transferable and diverse semantics during local training.

In this paper, we propose a \textbf{Fed}erated \textbf{D}istributionally \textbf{R}obust \textbf{A}lignment (FedDRA) framework, to learn transferable cross-modal alignment under data heterogeneity. Our key idea is to maximize cross-modal mutual information with distributional robustness. Specifically, to bridge the gap between the downstream testing domain and pre-training samples, we construct a set of distributions based on client distributions. We then employ a decentralized distributionally robust optimization method to iteratively improve the pre-trained model’s performance on this set. To alleviate the negative effect of over-fitting client-specific information, we maintain a global model to provide anchor representations for guiding local training and utilize a two-stage training schema to tune deep layers before updating the whole network.

Our contributions primarily focus on:
\begin{itemize}
  \item We for the first time tackle the problem of medical VLP under the federated setting by utilizing heterogeneous multi-modal datasets from different institutes. We conduct empirical studies to analyze the influence of data heterogeneity on federated multi-modal learning. 

  \item We propose FedDRA to address the data heterogeneity challenge in federated VLP to obtain transferable cross-modal alignment. It iteratively optimizes model performance on a distribution family and uses a two-stage global-guided local training strategy to reduce over-fitting on client-specific patterns. 

  \item Experiment results show the effectiveness of our method in learning multi-modal representations under the federated setting for various downstream tasks, including image-text retrieval, classification, and segmentation. 

\end{itemize}

\section{Related Work}
%\subsection{Medical Vision-Language Pre-training}
\textbf{Medical Vision-Language Pre-training. }Pre-training multi-modal models on large-scale datasets and then transferring learned knowledge to downstream tasks has become a popular approach to leverage diverse semantics contained in multi-modal unlabeled data ~\citep{li2022blip,bao2022vlmo,radford2021learning}. 
Current works aim to learn a shared latent space to connect the representations of each modality, leveraging a wide range of self-supervised learning methods, i.e., contrastive learning~\citep{radford2021learning, chen2020graph} and multi-modal fusion~\citep{li2021align, chen2022multi}. Medical multi-modal pre-training tasks are often conducted on vision-based datasets, especially vision-language pre-training. ~\citep{zhang2022contrastive} first utilizes an image-text contrastive loss to align visual and language representations. ~\citep{huang2021gloria} aligns fine-grained cross-modal representations~\citep{huang2021gloria} through a word-patch contrastive loss, and has improved the performance on fine-grained vision tasks. Furthermore, recent work~\citep{wang2022multi,bannur2023learning} incorporates medical domain knowledge to mitigate the misalignment during pre-training. %~\citep{hager2023best} also have considered aligning tabular data with images via self-supervised learning schemas, which has been extended to medical domain~\citep{du2024tip, yang2024vilref} to learn generalizable patterns shared within medical images and Electrical Health Records (EHR). 
However, almost all of the current methods still rely on large-scale pre-training datasets, which impede their adaptaion to modalities with limited training samples and deployment in real-world scenarios. 

\textbf{Heterogeneity in Self-Supervised Federated Learning. }Federated self-supervised learning aims to leverage diverse semantics in local unlabeled datasets in a decentralized and privacy-preserved way. One of the key challenges of federated learning is data heterogeneity~\citep{li2019fedmd,collins2021exploiting,li2021model}, and has been long discussed in the uni-modality scenarios. Typically, ~\citep{zhang2023federated,huang2022learn,li2022mocosfl} employ additional communications on local data representations to increase sample diversity. Such methods fail to protect data privacy, which is a vital concern in medical applications. On the other hand, ~\citep{zhuang2021collaborative, li2021model} utilizes server model to constrain the update of local models, ~\citep{yan2023label} utilizes the mask-autoencoder to handle heterogeneity, ~\citep{zhuang2022divergence,li2021model, han2022fedx} considers distillation-based methods yet ignores the direct modeling of cross-modal alignment. However, these uni-modal self-supervised learning methods have not accounted for the modality gap~\citep{zhang2024prototypical} between multi-modal data. While uni-modal self-supervised learning aims to learn robust features~\citep{radford2021learning}, multi-modal learning need also align input modalities to maximize the mutual information between their representations~\citep{su2023towards}. Recent advances~\cite {lu2023scaling} have verified that federated learning can be utilized to scale up the pre-training dataset. 
However, this work hasn't considered the harm of data heterogeneity issue~\citep{ghosh2019robust}. While the learned local models can be biased and over-reliance on spurious correlations~\citep{saab2022reducing} that are client-specific, distributionally robust optimization~\citep{deng2020distributionally} framework can alleviate these issues by optimizing the group-wise worst-case performance on given objective~\citep{liu2022distributionally}, thus this idea can be flexibly adapted to various of federated learning tasks~\cite {han2023distributionally,rehman2023dawa,capitani2024clusterfix}. 

\begin{table}

  \caption{Comparison between related works and our proposed FedDRA for federated vision-language pre-training. We organize the related works based on task settings and technical points:(1) Heterogeneous client datasets. (2) Multi-modal paired datasets. (3) Server-computation-free. (4) No communication on training data.(5) Introduce global constraints. (6) Distributionally robust framework.}
  \vspace{-10pt}
\resizebox{\textwidth}{!}{
  \centering
  \begin{tabular}{l llll ll}
    \toprule
    Related Work &\multicolumn{4}{c}{\textbf{Task Settings}} &\multicolumn{2}{c}{\textbf{Related Technical Points}}\\
    \cmidrule(l){2-5}  \cmidrule(l){6-7}
    & Hetero. & Multi-Modal & Server Comp.-Free & Feat. Commu.-Free & Global Const. & Dist. Robust\\
    \midrule
    ~\cite{zhuang2021collaborative} & $\checkmark$ & & $\checkmark$ & $\checkmark$  & $\checkmark$ & \\
    %~\cite{li2021model} & $\checkmark$ & & $\checkmark$ & $\checkmark$ & $\checkmark$ &\\
    %~\cite{han2022fedx} & $\checkmark$ & & $\checkmark$ & $\checkmark$ & $\checkmark$ & \\
    ~\cite{zhuang2022divergence} & $\checkmark$ & & $\checkmark$ & $\checkmark$ & & $\checkmark$\\
    ~\cite{zhang2023federated}& $\checkmark$ & & $\checkmark$ &  & $\checkmark$ & \\
    ~\cite{yan2023label} & $\checkmark$ & & & $\checkmark$  & &\\
    ~\cite{lu2023scaling} &  & $\checkmark$ & $\checkmark$ & $\checkmark$  & & \\
    \midrule
    Ours &$\checkmark$ & $\checkmark$ & $\checkmark$ & $\checkmark$ & $\checkmark$& $\checkmark$ \\
    \bottomrule
  \end{tabular}
}
\vspace{-10pt}
\label{diff}
\end{table}

\section{Problem Formulation}
\textbf{Formulation of Pre-training Dataset and Heterogeneity.} 
\label{sec data formulate}
In this paper, we consider the multi-modal datasets with two modalities $X,Y$, e.g., image and text modalities. Following~\citep{su2023towards}, we assume sample $x$ of modality $X$ and sample $y$ of modality $Y$ are generated from latent semantics through implicit mappings that are consistent across all clients. For instance, disease labels of a given X-ray image and radiology-report pair, are latent semantic variables that connect these two modalities. That's because these labels determine the pathology region of the radiology image and corresponding description in the diagnosis report. In federated learning, we consider $N$ clients, each has a own local dataset, forming a group $\mathcal{C}$. We assume that each client has a correspond data distribution $D_i, i \in [1, \dots, N]$, and data samples are given as $(x,y)\sim D_i$. In real-world scenarios, the distributions $D_i$ of local datasets vary across clients, introducing data heterogeneity that can negatively impact federated learning performance.

To obtain a generalizable model that performs well on testing domain, we often consider a virtual global dataset with data distribution $D_{\mathcal{T}}$~\citep{zhang2024eliminating}. In real-world setting, testing domains are often out-of-distribution (OOD), not limited to the pre-training local datasets. For example, a medical multi-modal model might be pre-trained on data from routine clinical practice and then transferred to tasks utilizing datasets collected during the COVID-19 pandemic. Therefore, we consider a family of global data domains that includes distribution shifts, which can be written in a form of uncertainty set: $Q^c: \{Q: D_f(Q \| D^{\mathcal{C}}) \leq \rho\}$, %\begin{equation}
%\label{uncertainty set}
%    Q^c: \{Q: D_f(Q \| D^{\mathcal{C}}) \leq \rho\}
%\end{equation}
where $D^{\mathcal{C}}$ is the distribution when entire data is grouped based on group $\mathcal{C}$, $D_f$ is the f-divergence of two distributions. $\rho\in R^+$ is the uncertainty radius, a larger $\rho$ introduces more unseen distributions.

\label{problem formulation}
\textbf{Federated Vision-Language Pre-training.} Given $N$ clients and their local datasets, federated learning aims to utilize the client dataset to train a generalizable model in a privacy-preserved way. It iteratively trains local models on the client side and aggregates (e.g., FedAvg strategy simply averages model parameters) them on the server. For each communication turn $r$, each client learns a local model through $E$ update steps, %\footnote{In this paper, we assume each client has the same number of iterations during local training in a communication turn.} 
and sends it to the server, and overwrites the local model with the aggregated model sent back from the server. Specifically, Federated Multi-Modal Pre-training aims to effectively leverage paired and unlabeled multi-modal data from local clients to learn a generalizable model $\hat{f}$. 

Multi-modal pre-training utilizes paired data from multiple modalities to learn model $\hat{f}$ that can well represent the samples. In this paper, we consider the pre-training task on image and text modalities. We focus on a classic schema in vision-language pre-training, where the model consists of feature encoders w.r.t. $X$ and $Y$ modalities, and both encoders project their inputs into a shared representation space $\mathcal{Z}$. For example, a good pre-trained model that can encode an image of a running dog and its text description "a photo of running dog" to a shared representation space $\mathcal{Z}$, which is called \textit{cross-modal alignment} in~\citep{castrejon2016learning,gao2024softclip}. Suppose the quality of the representation space of pre-trained models can be measured by a loss objective $\mathcal{R}: \mathbb{R}^{d\times d}\to \mathbb{R}$ (e.g., mutual information between representations $X$ and $Y$ with dimension $d$), federated multi-modal pre-training aims to minimize $\mathbb{E}_{(x,y))\sim D_{\mathcal{T}}}[\mathcal{R}(\hat{f}(x,y))]$ on the testing dataset $D_{\mathcal{T}}$. 

In federated setting, $\hat{f}$ is aggregated from $\hat{f_i}$, which are learned by minimizing $\mathbb{E}_{(x,y))\sim D_i}[\mathcal{R}(\hat{f_i}(x,y))]$ during local training. $\hat{f_i}$ may capture client-specific information that may not be generalizable across client dataset domains and will affect the performance of the aggregated $\hat{f}$ if local datasets are heterogeneous.

Table~\ref{diff} provides a comparison of the most similar previous works, highlighting the distinctions between their tasks and ours, as well as the technical differences between their approaches and ours.

\section{Method}
\begin{figure}
    \centering
    \includegraphics[width=1\linewidth]{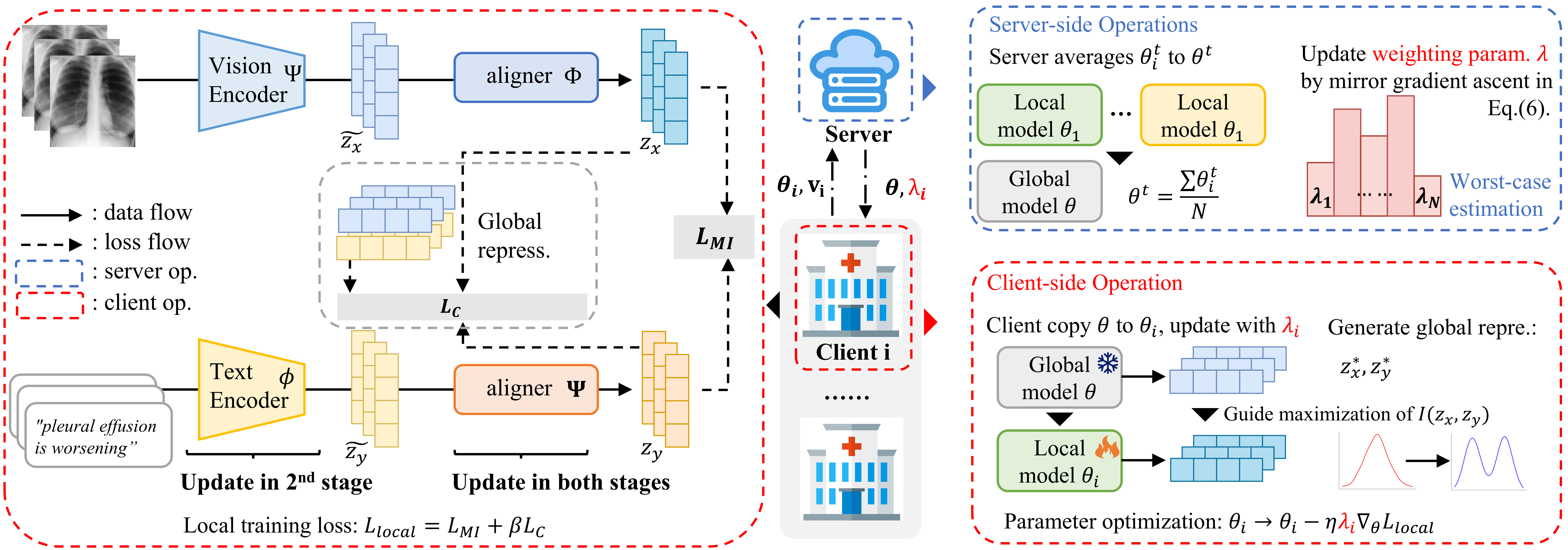}
    \caption{Our proposed FedDRA method follows a two-stage pre-training schema. Alignment modules are first trained in the first stage. Then we jointly update the whole model in the second stage. During local training, we use frozen copies of the server-aggregated models to obtain global representations $z^*_x$ and $z^*_y$, to regularize local training. After each round of local training, the server estimates the current worst-case distribution and updates the $\lambda_i$, which are then sent to each client to adjust the local update step-size.}
    \label{fig:method_overview}
    \vspace{-10pt}
\end{figure}

\subsection{Global Constrained Local Training Objective}
During local training, the pre-trained model would capture client-specific information that can not generalize to other data domains, as shown in Sec.~\ref{empirical finding}. Here, we will provide an in-depth analysis of this phenomenon and propose a global constraint term to alleviate it.

In classical vision-language pre-training setting, the vision-language model $f$ is composed of two encoders, $f_{\psi}: \mathcal{X} \to \mathcal{Z}$ for image modality $X$, and model $f_{\phi}: \mathcal{Y} \to \mathcal{Z}$ for text modality $Y$. Given an image-text paired data $(x,y)$, these models project the input to representations $z_x=f_{\psi}(x), z_y=f_{\phi}(y)$. The goal of vision-language pre-training is to learn a cross-modal alignment from unpaired data, and thus obtain a generalizable representation space, where image representation $z_x$ and text representations $z_y$ are well-aligned. It could be viewed as maximizing the mutual information of representations of the two modalities~\citep{su2023towards}. Therefore, we can measure the cross-modal alignment degree with a loss mutual-information-based loss objective $\mathcal{R}$, which is approximated by InfoNCE~\citep{liu2021learning,lu2024f} in this paper.

Current multi-modal pre-training methods often encourage the model to maximize mutual information $I(z_x, z_y)$ of the training pairs, neglecting the potential data heterogeneity problem in the federated learning scenarios. %For each client $i$, the latent variable $s$ is drawn from the distribution $D_i$ and generates $(x, y)$. As the distribution $D_i$ varies across clients, the observed distribution of training pairs $(x, y)$ of each client also varies. 
As the distribution $D_i$ varies across clients, each client dataset corresponds to a distinct optimal model $f_i$, which is induced from the distribution of the client dataset. For local training in client $i$, given that only $(x, y) \sim \mathcal{D}_i$ are available, the locally learned encoders $\hat{f_i}$ tends to move towards $f_{\psi_i}$ and $f_{\phi_i}$, and might capture some harmful client-specific information. In federated pre-training, the model is expected to capture patterns that are transferable across clients and potential testing domains, and client-specific information might harm the model's generalization ability. Therefore, it is crucial to explicitly force the model to learn generalizable knowledge. Previous methods such as FedAvg~\citep{mcmahan2017communication,lu2023scaling}, which do not account for this distinction, may result in learning biased local models $\hat{f_i}$, and diminish the generalization ability of the aggregated model $\hat{f}$.

Given the distribution \( D_{\mathcal{T}} \) of testing data domain, we aim to minimize the generalization error of the federally learned model on $D_{\mathcal{T}}$. Formally, let $\mathcal{F}_{\psi}\subseteq\{f_{\psi} : \mathcal{X} \rightarrow \mathcal{Z}\}$ to be a hypothesis space defined on input space of image modality $X$ and $\mathcal{F}_{\phi}\subseteq \{f_{\phi} : \mathcal{Y} \rightarrow  \mathcal{Z}\}$ of text modality $\mathcal{Y}$. Suppose encoders $\hat{f_{\psi}} \in \mathcal{F}_{\psi}$ and $\hat{f_{\phi}} \in \mathcal{F}_{\phi}$ are the federally learned models, and there exist optimal $f_{\psi_i}\in \mathcal{F}_{\psi}$ and $f_{\phi_i} \in \mathcal{F}_{\phi}$ for each data domain $D_i$. Denote $\epsilon_{g,h}(x)=\|g(x)-h(x)\|^2_2$ to be the error of two models $g(\cdot),h(\cdot)$ on data sample $x$, $\mathcal{L}$ as the InfoNCE loss. We have an upper bound of the generalization error $\mathcal{R}_T(\hat{f})=\mathbb{E}_{(x,y)\sim D_{\mathcal{T}}}[\mathcal{L}(\hat{f_{\psi}}(x), \hat{f_{\psi}}(y))]$.

\begin{proposition}
\label{error bound}
    Let $\{D_i, f_{\psi_i},f_{\phi_i}\}_{i=1}^N$ and $D_{\mathcal{T}}, f_{\psi_{\mathcal{T}}},f_{\phi_{\mathcal{T}}}$ be the distributions and optimal encoders for each client data domains and the testing domain, respectively. Given mixed weights $\{w_i\}_{i=1}^N$, $\sum_{i=1}^N w_i = 1$, $w_i \geq 0$, federally learned model $\hat{f}$, and temperature $\tau$ in InfoNCE loss. The generalization error $\mathcal{R}_T(\hat{f})$ follows:
\[
\mathcal{R}_T(\hat{f}) \leq   \mathbb{E}_{(x,y) \sim \mathcal{D}_{\mathcal{T}}} \alpha_i \cdot w_i\cdot \left[ \epsilon_{\hat{f_{\psi}}, \hat{f_{\psi_i}}}(x)+ \epsilon_{f_{\psi_i},\hat{f_{\psi_i}}}(x) + \epsilon_{\hat{f_{\phi}}, \hat{f_{\phi_i}}}(y)+ \epsilon_{f_{\phi_i},\hat{f_{\phi_i}}}(y) + C_i\right] .
\]
\end{proposition}
%where $C_i=8\tau log(bz)+\|f_{\psi_i}(x)-f_{\phi_i}(y)\|_2$ is a client-specific constant.
where $C_i, \alpha_i$ are client-specific constants.

Here, $\epsilon_{f_{\psi_i},\hat{f_{\psi_i}}}(x) $ and $\epsilon_{f_{\phi_i},\hat{f_{\phi_i}}}(x)$ measure the discrepancy between the locally trained models and the optimal models of the local data domain. These discrepancies are minimized during local training, leading the local models to learn client-specific information. Another two terms $\epsilon_{\hat{f_{\psi}},\hat{f_{\psi_i}}}(x)$ and $\epsilon_{\hat{f_{\phi}}, \hat{f_{\phi_i}}}(x)$ capture the discrepancy between the server-aggregated models and the locally trained models. Minimizing these terms can not only help reduce the upper bound of the generalization error, but also encourage the local models to learn patterns that generalize well to unseen testing data domains.

Motivated by this, we can directly optimize the terms $\epsilon_{\hat{f_{\psi}}, \hat{f_{\psi_i}}}(x)$ and $\epsilon_{\hat{f_{\phi}}, \hat{f_{\phi_i}}}(x)$ to encourage the models to learn generalizable features. By projecting the inputs $x$ and $y$ through $\hat{f_{\psi}}$ and $\hat{f_{\phi}}$, we obtain the global representations $z^*_x$ and $z^*_y$, respectively. Therefore, the constraint loss can be defined as $L_{C} = ||z_x - z^*_x||^2 + ||z_y - z^*_y||$. The total loss objective for local training could be:
\begin{equation}
\label{total loss}
    L_{local}=L_{MI}+\mu L_{C},
\end{equation}
where $\mu$ is the hyper-parameter that adjusts the constrained degree, $L_{MI}$ is the pre-training loss term based on infoNCE loss (e.g., image-text contrastive loss).

\subsection{Two-Stage Alignment For Mitigating the Deeper Layer Bias}

Furthermore, as discussed in Sec.~\ref{empirical finding}, deeper layers that contain biased client-specific information can impede pre-trained model to learn generalizable representations. This observation is similar to the findings in the supervised federated learning domain~\citep{legate2024guiding}, where training from a better initialized last layer, can less capture biased information in client local datasets. Motivated by this, we model the deep layers of the encoding functions of modality $X,Y$ as alignment modules $f_{\Psi},f_{\Phi}$, and aim to obtain generalizable alignment modules. In practice, we add additional blocks as the alignment module for simplicity, instead of dividing each encoder to two separate parts. For a data flow of an input pair $(x,y)$, the image encoder and text encoder $f_{\psi},f_{\phi}$ firstly take $x,y$ to obtain intermidiate features $\tilde{z}_x,\tilde{z}_y$, respectively. Then the corresponding alignment modules $f_{\Psi},f_{\Phi}$ will project $\tilde{z}_x,\tilde{z}_y$ to aligned final representation $z_x,z_y$.

To mitigate the negative impact of the biased alignment module on the generalization ability of the pre-trained model, we first train generalizable alignment modules $f_{\Psi}$ and $f_{\Phi}$, and then use them to enhance the training of the feature encoders $f_{\psi}$ and $f_{\phi}$. To encourage alignment modules learn to extract general features, in the first stage, we train them with frozen feature encoders using the learning objective Eq.(\ref{total loss}). Since these feature encoders are less biased from client-specific information, alignment modules are encouraged to learn unbiased mappings from $\tilde{z}_y$ to $z_y$ and $\tilde{z}_x$ to $z_x$. Then in the second stage, we train both the alignment module and feature encoders to enhance their capability of extracting medical features, with the same learning objective as the first stage. The complete pipeline is illustrated in Fig.~\ref{fig:method_overview}.

\begin{figure}[h!]

    \centering
    \begin{minipage}[t]{0.52\textwidth} % 将左边的 minipage 宽度设为 0.55
        \begin{algorithm}[H]
            \footnotesize
            \caption{FedDRA for Federated V-L Pre-training}
            \begin{algorithmic}[1]
                \State \textbf{Input:} Init. encoder param. $\theta^{(0)}$; Init. alignment module param. $\Theta^{(0)}$; Client weight $\{w_{i}^{(0)}\}_{i=1}^N$; Uncertainty set radius $\rho$; Client datasets $\{D_i\}_{i=1}^{N}$; Local iteration steps $E$.
                \For{$r = 0, \ldots, R-1$}
                
                \State Server broadcasts $\Theta^{(rE)},\theta^{(rE)}, \theta^{(rE)*}, \{w_{i}^{(r)}\}_{i=1}^N$
                \For{client $i = 1, \ldots, N$}
                \State Utilize Algorithm 2 on $D_{i}$ to get $ \Theta_{i}^{(r+1)E}$
                \EndFor
                \State Server computes: $\Theta^{(r+1)E} = \frac{1}{N} \sum_{i=1}^{N} \Theta_{i}^{(r+1)E}$
                \State Server broadcasts $\Theta^{(r+1)E}$
                \For{client $i = 1, \ldots, N$}
                \State Utilize Algorithm 3 on $D_{i}$ to get $\theta_{i}^{(r+1)E}$
                \EndFor
                \State Server computes: $\theta^{((r+1)E)} = \frac{1}{N} \sum_{i=1}^{N} \theta_{i}^{((r+1)E)}$
                \For{client $i = 1, \ldots, N$}
                \State Compute loss $v_{i}^{(r+1)}$ 
                \State of model $(\Theta_{i}^{(r+1)E};\theta_{i}^{(r+1)E})$ on $D_{i}$
                \EndFor
                \For{client $i = 1, \ldots, N$}
                \State ${w_{i}^{(r+1)}} = Proj_{\mathcal{Q}{i}}\left( {w_{i}^{(r)}e^{\gamma v_{i}^{(r+1)}}}/\sum_{i=1}^{N}{w_{i}^{(r)}e^{\gamma v_{i}^{(r+1)}}} , \rho \right)$
                \EndFor
                \EndFor
                
                \State \textbf{return} ($\Theta^{(RE)}$; $\theta^{(RE)}$)
            \end{algorithmic}
        \end{algorithm}
    \end{minipage}
    \hfill
    \begin{minipage}[t]{0.44\textwidth} % 将右边的 minipage 宽度设为 0.4

        \begin{algorithm}[H]
            \footnotesize
            \caption{1st Stage Training}
            \begin{algorithmic}[1]
                \State \textbf{Input:} Learning rate $\eta$;Update stepsize $\gamma$; Local iteration steps $E$; Param. $\theta^{rE}$, $\Theta^{rE}$; Weight $w_i$
                \State Set $(\theta_{i}^{rE};\Theta_{i}^{rE}) = (\theta^{rE};\Theta^{rE})$
                \State Set $(\theta_{i}^{rE*};\Theta_{i}^{rE*}) = (\theta^{rE};\Theta^{rE})$
                \For{$t = rE, \ldots, (r + 1)E - 1$}
                \State Sample $\xi_{i}^{t}$ from $D_i$ uniformly
                \State $\theta_{i}^{t+1*} =\theta_{i}^{t*} - $
                \State ${w_{c}^{r+1}} \eta\nabla_{\Theta}L_{dro}(\Theta_{i}^{t};\theta_{i}^{t}; \Theta^{(r+1)E*};\theta^{(r+1)E*};\xi_{i}^{t})$
                \EndFor
            \end{algorithmic}
        \end{algorithm}

        \vspace{0.2cm} 
        
        \begin{algorithm}[H]
            \footnotesize
            \caption{2nd Stage Training}
            \begin{algorithmic}[1]
                \State \textbf{Input:} Learning rate $\eta$; Update stepsize $\gamma$; Local iteration steps $E$; Param. $\theta^{rE},\Theta^{(r+1)E}$; Weight $w_i$ 
                \State Set $(\theta_{i}^{rE};\Theta_{i}^{(r+1)E}) = (\theta^{rE};\Theta^{(r+1)E})$
                \State Set $(\theta_{i}^{rE*};\Theta_{i}^{(r+1)E*}) = (\theta^{rE};\Theta^{(r+1)E})$
                \For{$t = rE, \ldots, (r + 1)E - 1$}
                \State Sample $\xi_{i}^{(t)}$ from $D_i$ uniformly
                \State $(\theta_{i}^{t+1};\Theta_{i}^{t+1}) =(\theta_{i}^{t};\Theta_{i}^{t}) -{w_{c}^{r+1}}$
                \State $\eta\nabla_{(\theta_{i};\Theta_{i})}L_{local}(\theta_{i}^{t};\Theta_{i}^{t}; \Theta^{(r+1)E*};\theta^{(r+1)E*};\xi_{i}^{t})$
                \EndFor
                \State The client sends $\theta_{i}^{(r+1)E}, v_{c}^{r}$ to the server
            \end{algorithmic}
        \end{algorithm}

    \end{minipage}
\label{algorithm_all}
\vspace{-10pt}
\end{figure}

\subsection{Learning Robust Cross-Modal Alignment via Distributionally Robust Optimization}
In real-world scenarios, $\mathcal{D}_{\mathcal{T}}$ is unknown during pre-training and is typically out-of-distribution. A common approach to address this issue is to assume the distribution of the testing data domain is near the distribution of overall training data~\citep{rahimian2019distributionally,levy2020large}, and construct a set \( \mathcal{Q}^{\mathcal{C}} \) that covers potential testing distributions. By optimizing the model's parameter $\theta$ over this set with the loss objective $\mathcal{R}(\theta) := \sup_{Q \in \mathcal{Q}^{\mathcal{C}}} \left\{ \mathbb{E}_{(x,y) \sim Q} \left[ L_{\text{local}}(\theta, x, y) \right] \right\}$, we can pre-train a model that generalizes well to the whole set of potential testing distributions. Here, the maximum loss implies a worst-case distribution in $\mathcal{Q}^{\mathcal{C}}$, where the pre-trained model performs the worst in aligning the two modalities.

Inspired by this, we aim to optimize the loss objective on the worst-case distribution, and introduce the Distributionally-Robust-Optimization (DRO) to our federated multi-modal pre-training task. DRO first construct a family of testing distributions $\mathcal{Q}^{\mathcal{C}}$ as shown in Sec.~\ref{sec data formulate}, and optimize the model's performance on the worst-case distribution, where the model performs the poorest among distributions in $\mathcal{Q}^{\mathcal{C}}$. However, during federated learning, the server has no access to the distribution of the entire data. Motivated by~\citep{zhang2023federated}, we introduce a de-centralized form of the DRO problem. The optimization object could be written as:
\begin{equation}
\label{dro}
\sup_{\lambda \in \Delta_{|\mathcal{C}|-1}} \left\{ R(\theta, \lambda) := \sum_{i} \lambda_i R_i(\theta) \, \text{s.t.} \, D_f(|\mathcal{C}| \cdot \lambda || (1,1,\dots,1)) \leq \rho \right\},
\end{equation}
where \( R_i(\theta) := \mathbb{E}_{(x,y) \sim D_i}[L_{local}(\theta; (x,y))] \) is the empirical risk on client data $D_i$, $\rho$ is the uncertain radius as mentioned in Sec.~\ref{sec data formulate}. 

Then, we can optimize Eq.~\ref{dro} by alternatively optimize the weights $\lambda$ and model parameters $\theta$. Specifically, we optimize the parameter $\theta_i$ of local models on each iteration $t$ by $\theta_i^{(t+1)} = \theta_i^{(t)} - \eta \lambda_i^{t} \nabla_{\theta} L_{local}$, where $\eta$ is the learning rate. Following the mirror gradient ascent of weight proposed in~\citep{zhang2023federated}, we update $\lambda$ with $\lambda^{t+1} = \frac{\lambda_i^{t} e^{\gamma v_i^{t}}}{\sum_{i=1}^{|\mathcal{C}|} \lambda_j^{t} e^{\gamma v_j^{t}}}$. Then we compute \( \lambda^{t+1} \) by projecting \( \tilde{\lambda}^{t+1} \) into the set \( \{ \lambda : D_f(|\mathcal{C}|) \cdot \|\lambda\| \|(1,1,\dots,1)\| \leq \rho \} \) to fit the constraints of the uncertainty set. In practice, we update $\lambda$ after the local training of each communication turn.

We apply the proposed DRO in both the first stage of training the alignment module and the second stage of training the feature encoder. In both stages, we use the same objective, \( L_{local} \), to encourage the model to learn generalizable information and mitigate the impact of data heterogeneity on maximizing the mutual information \( I(z_x, z_y) \). The key difference between the two stages is, in the first stage, the optimization target \( \theta \) in Eq.~\ref{dro} corresponds to the parameters of the alignment modules \( \Phi \) and \( \Psi \), whereas in the second stage, \( \theta \) represents the parameters of the feature encoders \( \phi,\psi \) and alignment modules \( \Phi, \Psi \). The pseudo-code of the whole algorithm can be seen in Algorithm 1.

\section{Experiment}
\label{experiment}
\subsection{Experiment Setting}
\label{exp setting}
We focus on adapting medical vision-language pre-training methods to heterogeneous federated learning settings. We employ the framework of image-text contrastive learning with two modality-specific encoders, a fundamental design in multi-modal pre-training. We use vision-language pre-training tasks on Chest X-ray datasets and ophthalmology image datasets to evaluate the effectiveness of our FedDRA method.

\subsubsection{Experiment Set-up of Pre-training on Chest X-Ray datasets}
\textbf{Pre-training setup.} Following~\citep{wang2022multi}, we utilize the MIMIC-CXR~\citep{bigolin2022reflacx} dataset for medical vision-language pre-training. Following~\citep{yan2023label}, we employ the Latent Dirichlet Allocation (LDA)~\citep{blei2003latent} to divide the MIMIC-CXR dataset based on disease labels to construct 5 heterogeneous client datasets. We set the heterogeneity degree in the LDA algorithm to be 1. Each divided dataset consists of train splits and test splits based on the notation of the MIMIC-CXR. We use the train split for pre-training, and test split to evaluate pre-trained model's image-text retrieval performance. Here, we only divide the raw data into 5 subsets, because vision-language pre-training requires a large batchsize and is data-consuming, thus we need to guarantee each client has $10k$ to $50k$ paired data.

For main experiments, we set the number of communication turns to 25, and randomly sample 50 batches of data for local training at each turn. Here, we choose a relatively small number of communication turns compared to classical supervised federated learning. That's because VLP needs large local optimization steps per turn to extract cross-modal alignment.

\textbf{Downstream tasks.} Following~\citep{wang2022multi}, we conduct the following downstream tasks to evaluate the transferability and generalization ability of the pre-trained model. \textbf{(1) Few-shot classification. }We test their performance on multiple image classification benchmarks RSNA Pneumonia Detection (RSNA)~\citep{shih2019augmenting}, and Covidx~\citep{wang2020covid}. We fine-tune our pre-trained model with an additional linear layer on 1\%, 10\% percentage of the training dataset, and evaluate the classification accuracy. \textbf{(2) Medical image segmentation.} We conduct medical image segmentation experiments on the RSNA~\citep{wang2020covid} benchmark. We freeze the encoder and fine-tune a U-Net decoder using 1\%, 10\% of the training data, and then use the Dice score for evaluation. The datasets we have used for the fine-tuning based tasks are out-of-distribution, so that we can evaluate the transferability of the pre-trained model. \textbf{(3) Image retrieval.} We utilize the test splits of client datasets for evaluation, these datasets are unseen in pre-training, and can be viewed as in-domain samples. We report the top-1 recall accuracy and top-5 recall accuracy.

\subsubsection{Experiment Set-Up of Pre-training on Ophthalmology Datasets}
\textbf{Pre-training setup.} We conduct vision-language multi-modal pre-training using retinal image datasets from different institutes to simulate a more real-world setting. These retinal datasets are from different institutions of low-income and high-income countries, and are highly heterogeneous real-world scenes. Specifically, we utilize MESSIDOR~\citep{decenciere2014feedback} from France and BRSET~\citep{nakayama2023brazilian} from Brazil as pre-training datasets, and assign them to two clients. These datasets include both images and tabular EHR records indicating Diabetic Retinopathy (DR) status and edema risk. For implementation, we transform these tabular data into text captions.

\textbf{Downstream tasks} We evaluate the transferability of the models on few-shot classification tasks using the MBRSET~\citep{nakayamambrset} dataset. Unlike the pre-training datasets, MBRSET was collected by portable devices, resulting in a significant distribution shift. We perform few-shot classification tasks on diabetic retinopathy and edema status using this dataset. We fine-tune the model with an additional linear layer on $10\%$, $20\%$ and $100\%$ of the training data, and report classification accuracies.

\subsubsection{Backbones and Baselines} 
We focus on enabling medical multi-modal pre-training methods to be applied to heterogeneous federated learning scenes. We have considered the generalization ability of our method on different backbone VLP methods, and adopted contrastive-learning-based methods: simple language-image contrastive alignment (ConVIRT)~\citep{zhang2022contrastive,radford2021learning}, global-local language-image contrastive alignment(GLoRIA)~\citep{huang2021gloria}, and Multi-Granularity Cross-modal Alignment (MGCA)~\citep{wang2022multi}. All of the loss objectives of these pre-training methods contain a contrastive loss term, which act as the infoNCE loss to maximize the mutual information between two modalities. And we take this loss term for computing the client weights in the DRO part.

For baseline federated learning strategies, we have adapted FedMAE~\citep{yan2023label}, FedEMA~\citep{zhuang2022divergence}, FedMOON~\citep{li2021model}, FedX~\citep{han2022fedx}, FedU~\citep{zhuang2021collaborative}, FedLDAWA~\citep{rehman2023dawa} for comparison. These are self-supervised learning methods in the federated learning domain which also focus on tackling the data heterogeneity. For basic federated learning baselines, we consider simple averaging (FedAvg)~\citep{mcmahan2017communication}, decentralized training, and centralized training. For baselines pre-trained in Local strategy, we report the averaged performance of the local models.

For fair comparisons, we re-implemented all baseline methods using the same backbones. To adapt uni-modal self-supervised learning baselines to our setting, we added an image-text contrastive loss, applying the same hyperparameters as in our method for consistency. We use ViT-base~\citep{dosovitskiy2020image} as the vision encoder and Bert-base~\citep{devlin2018bert} as the text encoder, with input pre-processing following~\citep{wang2022multi}. Additionally, we employ an extra transformer block from ViT-base and Bert-base as the alignment module for vision and language, respectively.

\begin{figure}
    \centering
    \includegraphics[width=1\linewidth]{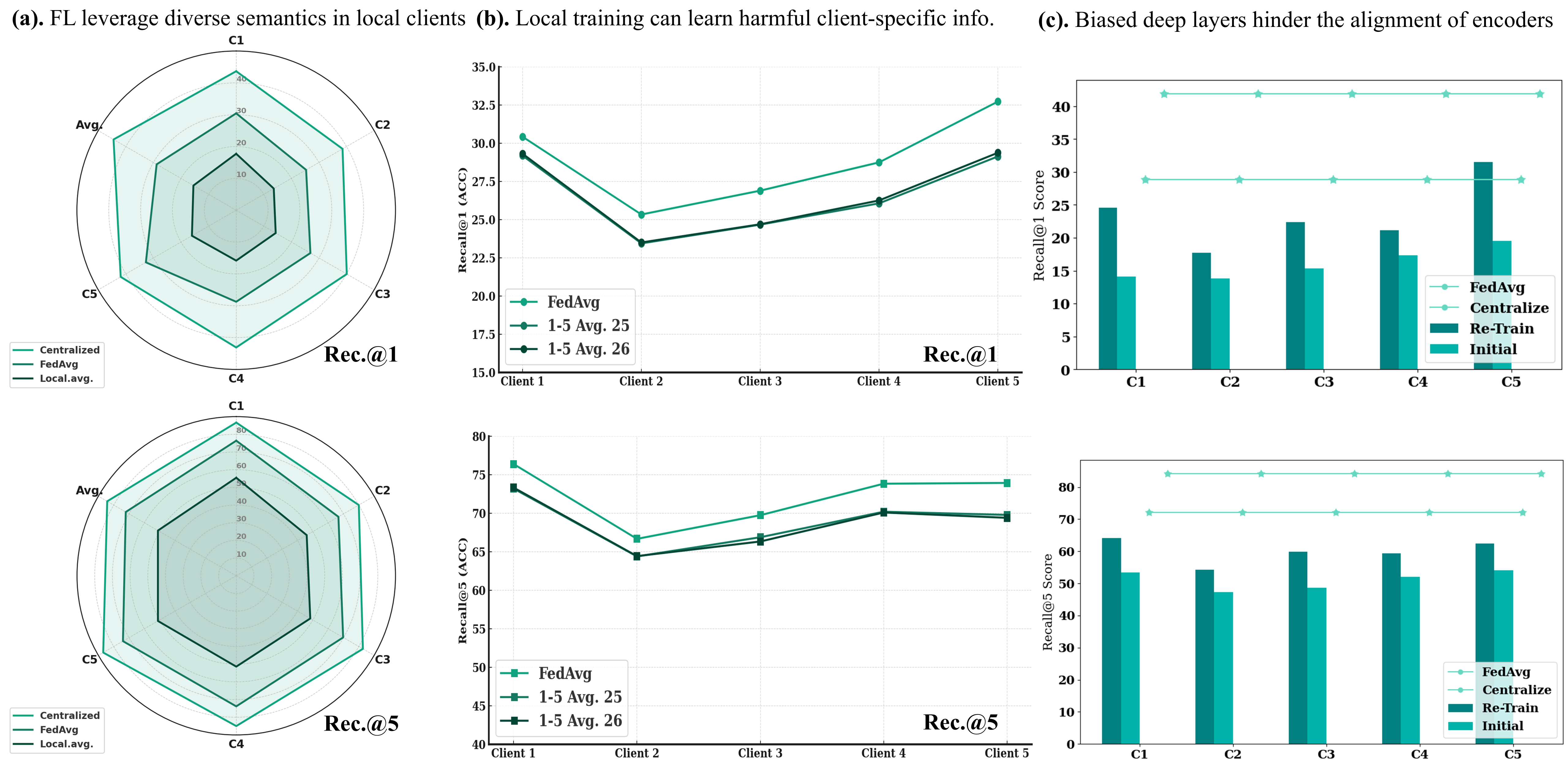}
    \caption{(a). The comparison of retrieval acc. on each client denoted as $\{C_i\}_{i=1}^5$, of centralized, FedAvg, and averaged acc. of decentralized pre-trained models. (b). The performance of the server model after 25 com. turns and the averaged performance of corresponding client models after 25 and 26 com. turns, on each client denoted as $\{\text{Client}_i\}_{i=1}^5$. (c). The averaged acc. on each client. We show the acc. of centralized and FedAvg pre-trained baselines, and de-centralized pre-trained models shown as $\{C_i\}_{i=1}^5$ retrained on the union of training splits of client datasets.}
    \label{fig:empirical findings}
    \vspace{-15pt}
\end{figure}
\subsection{Empirical Finding}
\label{empirical finding}
In this section, we will demonstrate our key empirical findings for federated multi-modal pre-training under heterogeneous client datasets, which actually motivates us to propose our FedDRA. We conduct experiments on the image-text retrieval task, which reflects the ability to maximize the mutual information and learn cross-modal alignment. In the following studies, we mainly compare performances of naive federated (FedAvg) pre-trained model, decentralized pre-trained model, and centralized pre-trained model. %Detailed results can be found in Table~\ref{tab:image_retrieval} in the appendix.

\textbf{Federated learning enhances pre-training by leveraging more samples in a privacy-preserved manner, while data heterogeneity can affect the effectiveness of the FedAvg.} Figure~\ref{fig:empirical findings}(a) presents the retrieval accuracies of the models under consideration. Despite the heterogeneity of local datasets, the FedAvg strategy significantly outperforms the decentralized pre-training approach. However, the centralized pre-trained model remains an upper bound, indicating substantial room for improvement. %Additionally, we investigate whether updating the aggregated server model in FedAvg leads to distortions in the learned representation space.

\textbf{Local training can learn harmful client-specific information, degrading the performance of the pre-trained model.} After several communication turns, re-training the aggregated server model on local datasets may lead to a performance drop, as shown in Figure~\ref{fig:empirical findings}(b). We considered a set of models retrained on a server model with local datasets respectively. The server model is learned through 25 communication turns. Compared to the starting server model, the averaged accuracy of local retrained models is significantly lower. This degradation may be because local training focus on learning domain-specific information in the late communication rounds, which would affect the aggregated model's overall performance.

\textbf{Decentralized pre-trained deeper layers can hinder the learning of a generalizable feature extractor.} We re-trained the first four shallow layers of the decentralized pre-trained model on the combined local datasets. While this led to some performance improvements, a significant gap still remains compared to the FedAvg pre-trained baselines, as shown in Figure~\ref{fig:empirical findings}(c). This gap indicates that the biased frozen deep layers prevent the model from learning more diverse semantics from the combined dataset. We hypothesize that these deep layers may contain biased, client-specific information, which obstructs the cross-modal alignment process. Our findings align with observations~\citep{legate2024guiding} in supervised federated learning.

Overall, from empirical findings, we conclude that federated multimodal pre-training is sensitive to data heterogeneity, and simply averaging local model weights does not solve this issue essentially. Furthermore, performance is closely tied to the generalization ability of the final layers in pre-trained models. Thus, we are motivated to propose our method.
\begin{table*}[h] % The 't' means "top of the page"
\centering % This centers the table

\caption{Downstream task performance. We report the few-shot classification accuracy on Covid and RSNA, the in-domain retrieval accuracy, and the Dice score for segmentation on RSNA. 
}
\resizebox{\textwidth}{!}{
\begin{tabular}{@{} ll ll ll  ll  llll }
\toprule
\textbf{Strategy} & \textbf{Backbone}&\multicolumn{2}{c}{\textbf{RSNA (cls.)}} &\multicolumn{2}{c}{\textbf{Covid (cls.)}} &\multicolumn{2}{c}{\textbf{RSNA (seg.)}}  &\multicolumn{4}{c}{\textbf{In-domain Image-Text Retrieval}}\\
\cmidrule(l){3-4}  \cmidrule(l){5-6}  \cmidrule(l){7-8}  \cmidrule(l){9-12}  
 & & \textbf{$1\%$} & \textbf{$10\%$} & \textbf{$1\%$} & \textbf{$10\%$}  & \textbf{$1\%$} & \textbf{$10\%$}  & \textbf{Rec.@1} & \textbf{Rec.@5} & \textbf{Wst.@1} & \textbf{Wst.@5}\\ 
\midrule

FedEMA & ConVIRT & 82.8 & 83.1 & \underline{79.2} & 86.5 & \underline{70.9} & \underline{73.6} & 24.0 & 67.0 & 21.9 & 62.4 \\
FedMOON & ConVIRT & 82.5 & 83.2 & 77.8 & \underline{89.2} & 69.0 & 71.3 & 27.8 & 70.9 & 25.3 & \underline{67.2} \\
FedAvg & ConVIRT & \underline{83.1} & \underline{83.3} & 78.0 & 88.5 & 69.6 & 71.5 & \underline{28.8} & \underline{72.1} & \underline{25.3} & 66.7 \\
\rowcolor{gray!30}
FedDRA (Ours)& ConVIRT & \textbf{83.2} & \textbf{83.7} & \textbf{81.0} & \textbf{90.3} & \textbf{71.7} & \textbf{74.1} & \textbf{30.2} & \textbf{73.2} & \textbf{27.0} & \textbf{68.9}\\
\midrule
FedX & GLoRIA & 82.7 & 83.4 & 78.3 & 88.5 & 71.0 & 72.1 & 28.5 & 72.2 & 25.9 & 68.0\\
FedU & GLoRIA & 83.0 & \underline{83.5} & \underline{78.7} & \underline{89.3} & 71.2 & \underline{72.6} & 29.2 & 73.0 & 27.6 & 69.5\\
FedAvg & GLoRIA & \underline{83.2} & 83.3 & 77.5 & \underline{89.0} & 71.4 & 72.4 & \underline{29.9} & \underline{73.8} & \underline{27.8} & \underline{69.5} \\
\rowcolor{gray!30}
FedDRA (Ours) & GLoRIA & \textbf{83.6} & \textbf{84.1} & \textbf{79.4} & \textbf{89.8} & \textbf{72.0} & \textbf{73.2} & \textbf{31.1} & \textbf{74.3} & \textbf{28.2} & \textbf{70.2}\\
\midrule
FedLDAWA & MGCA & 82.4 & \underline{83.5} & \underline{78.1} & \underline{88.5} & \underline{70.4} & \underline{72.6} & 29.0 & 73.5 & \underline{27.0} & 68.9\\
FedAvg & MGCA & \underline{82.6} & 83.5 & 75.8 & 88.2 & 70.1 & 71.4 & \underline{29.3} & \underline{73.7} & 26.8 & \underline{70.4}\\
\rowcolor{gray!30}
FedDRA (Ours) & MGCA& \textbf{83.1} & \textbf{83.8} & \textbf{79.3} & \textbf{89.1} & \textbf{71.0} & \textbf{72.8} & \textbf{29.8} & \textbf{74.1} & \textbf{27.4} & \textbf{70.6}\\
\bottomrule
\end{tabular}
}
\label{tab:downstream_main}
\vspace{-10pt}
\end{table*}

\begin{table}[h] % The 't' means "top of the page"
\centering % This centers the table

\caption{We have conducted ablation experiments to verify the effectiveness of our key technical designs. We report the few-shot classification accuracy and the retrieval accuracy.}

\resizebox{\linewidth}{!}{
\begin{tabular}{@{} lll llll llll}
\toprule
\textbf{Two-stage} & \textbf{Global Constraint} & \textbf{DRO-Weighing} &\multicolumn{2}{c}{\textbf{Covid (cls.)}} &\multicolumn{2}{c}{\textbf{RSNA (cls.)}} &\multicolumn{4}{c}{\textbf{In-domain Image-Text Retrieval}}\\
\cmidrule(l){4-7} \cmidrule(l){8-11} 
 & & &\textbf{$1\%$} & \textbf{$10\%$}  &\textbf{$1\%$} & \textbf{$10\%$} &\textbf{Rec.@1} & \textbf{Rec.@5}&\textbf{Wst.@1} & \textbf{Wst.@5}\\ 
\midrule
& $\checkmark$ & $\checkmark$ & 83.0 & 83.4 & 80.5 & 89.6 & 29.4 & 72.7 & 26.2 & 68.1\\
$\checkmark$& & $\checkmark$ & 82.8 & 83.0 & 79.8 & 88.6 & 28.3 & 71.9 & 26.0 & 67.9\\
 $\checkmark$ & $\checkmark$ & & 82.5 & 82.9 & 80.2 & 89.2 & 29.7 & 72.8 & 25.6 & 67.3\\
\midrule
$\checkmark$ & $\checkmark$ & $\checkmark$ & \textbf{83.2} & \textbf{83.7} & \textbf{81.0} & \textbf{90.3} & \textbf{30.2} & \textbf{73.2} & \textbf{27.0} & \textbf{68.9}\\
% ... Your other rows here
\bottomrule
\end{tabular}
}
\label{tab:ablation}
\vspace{-10pt}
\end{table}

\subsection{Main Results}
\textbf{Our method learns robust and enriched cross-modal alignment and has better transferability.} Table~\ref{tab:downstream_main} has shown results of downstream tasks, here we utilize the ConVIRT as the backbone pre-training method. In the image-text retrievel task, both average and worst-client accuracies of our method are higher than baseline's, which means our model can capture more robust cross-client features. In the few-shot classification and segmentation, our method beats other baseline strategies on each task, which demonstrate the higher generalization ability of the representation space learned by our method. 

%\textbf{Previous single-modality methods cannot be adapted directly for multi-modal data.} 
Table~\ref{tab:downstream_main} has shown the performance of adapted self-supervised federated learning methods which focus on single-modality. From the results, we observe that baselines have shown better transferability on visual downstream tasks, compared to the naive FedAvg strategy. However, in the multi-modal retrieval task, our method beats these baselines by a large margin, which indicates that previous single-modality methods cannot be easily adapted directly for multi-modal data. Furthermore, FedAvg is a strong baseline in multi-modal retrieval tasks compared to other adapted methods, as we observed in the experiments. We conjecture that's because FedAvg only focuses on maximizing the in-domain mutual information, and doesn't introduce additional loss terms which would hurt the learning of enriched cross-modal alignment. However, this may lead to lower generalization ability on few-shot downstream tasks as discussed before.

\begin{minipage}{0.58\textwidth} % Adjust width as needed
\centering
\captionof{table}{Downstream task performance. We report the few-shot classification accuracy of Diabetic Retinopathy and Edema Risk classification tasks on the MBRSET dataset.}
\resizebox{\textwidth}{!}{
\begin{tabular}{@{} l llllll }
\toprule
\textbf{Strategy}  &\multicolumn{3}{c}{\textbf{Diabetic Retinopathy (cls.)}} &\multicolumn{3}{c}{\textbf{Risk of Edema (cls.)}} \\
\cmidrule(l){2-4}  \cmidrule(l){5-7}  
  & \textbf{$10\%$} & \textbf{$20\%$} & \textbf{$100\%$} & \textbf{$10\%$} & \textbf{$20\%$} & \textbf{$100\%$}\\ 
\midrule
Decentralized  & 78.8 & 79.7 & 81.1 & 91.5 & 92.5 & 93.8\\
FedAvg  & 79.4 & 80.2 & \underline{82.3} & 92.8 & \underline{93.6} & 94.2\\
FedMAE~\citep{yan2023label}  & 79.2 & 80.3 & 82.0 & 92.4 & 93.3 & 94.0\\
FedX~\citep{han2022fedx}  &79.5 & 80.1 & 81.6 & \underline{93.0} & 93.5 & \underline{94.3}\\
FedU~\citep{zhuang2021collaborative}  & \underline{79.7} & \underline{80.5} & 81.7 & 92.8 & 93.4 & 94.1\\
FedDRA (Ours)  & \textbf{80.6} & \textbf{81.5} & \textbf{83.1} & \textbf{93.4} & \textbf{94.1} & \textbf{94.9}\\
\midrule
%\rowcolor{gray!30}
FedGlobal  & 81.9 & 82.6 & 84.0 & 94.2 & 94.7 & 95.8\\
\bottomrule
\end{tabular}
}
\label{tab:retinal_main}
\end{minipage}
\hfill
\begin{minipage}{0.38\textwidth} % Adjust width as needed
\textbf{Our method can be transferred to multiple multi-modal pre-training methods.} Table~\ref{tab:downstream_main} shows the downstream task performance of the MGCA and GLoRIA backbone pre-training methods when combined with our strategy. Our method has successfully adapted MGCA and GLoRIA to the heterogeneous federated multi-modal pre-training scenario, as demonstrated by the significant improvement in classification and segmentation tasks.
\end{minipage}

\subsection{Analysis Experiments}

\begin{figure}[htp] 
    \centering
    \includegraphics[width=\textwidth]{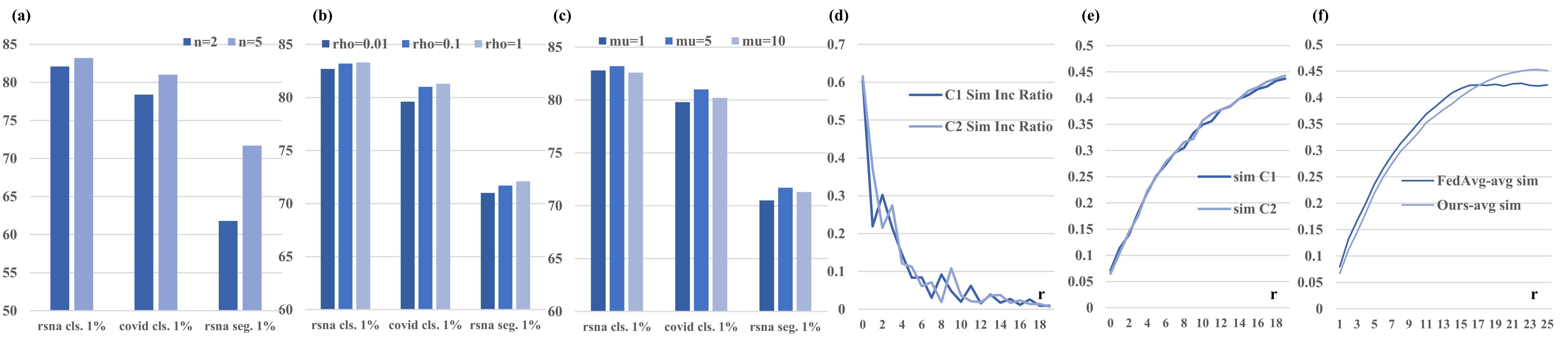} % Adjust the path and width as needed
    \caption{ (a) Analysis study on number of clients. (b) Analysis study on uncertainty radius. (c) Analysis study on global constraint degree. (d) Improvement of image-text embedding similarity per.commu turn. (e) Image-text embedding similarity curve of our method. (f) Averaged similarity curves.
    }
    \label{fig:all_ana}
\end{figure}

\textbf{The two-stage pre-training strategy and global constraints can enhance the learning of cross-modal alignement.} We remove the global constraint loss $L_{C}$ of our method, and compare the pre-trained model's performances with those of the original version. As shown in Table~\ref{tab:ablation}, the downstream performance, particularly the image-text retrieval accuracy, is significantly lower in the modified version. Similarly, we remove the first-stage pre-training on alignment modules to verify the role of the two-stage pre-training strategy. We have found that the first stage pre-training can help learn better cross-modal alignment and achieve high image-text retrieval accuracies, as shown in Table~\ref{tab:ablation}.

\textbf{DRO weighing can reduce the domain gap and improve the downstream performances.} We remove the DRO weighing part and compare the performances of the model pre-trained with the original method. As shown in Table~\ref{tab:ablation}, removing the DRO-weighing part leads to a large performance drop in few-shot classification performances, thus adding DRO-weighing component can improve the transferability of the pre-trained model. The client-wise worst accuracies of the original model are much higher than those of the modified version. This means DRO has succesfully bridged the gap between local training data and downstream dataset, by optimizing model performance on the constructed uncertain set of distributions.

\begin{table*}[h] % The 't' means "top of the page"
\centering % This centers the table

\caption{Results on client datasets with different heterogeneity degrees, across different pre-training methods.}
\resizebox{\linewidth}{!}{
\begin{tabular}{@{} ll l llllll | l llllll}
\toprule
\textbf{Strategy}  & \textbf{$\alpha$} & \textbf{Backbone} &\multicolumn{2}{c}{\textbf{RSNA (cls.)}} &\multicolumn{2}{c}{\textbf{Covid (cls.)}} &\multicolumn{2}{c}{\textbf{RSNA (seg.)}} & \textbf{Backbone} &\multicolumn{2}{c}{\textbf{RSNA (cls.)}}  &\multicolumn{2}{c}{\textbf{Covid (cls.)}} &\multicolumn{2}{c}{\textbf{RSNA (seg.)}}\\
\cmidrule(l){4-5}  \cmidrule(l){6-7} \cmidrule(l){8-9} \cmidrule(l){11-12}  \cmidrule(l){13-14}  \cmidrule(l){15-16} 
 & & & \textbf{$1\%$} & \textbf{$10\%$} & \textbf{$1\%$} & \textbf{$10\%$}& \textbf{$1\%$} & \textbf{$10\%$}& & \textbf{$1\%$} & \textbf{$10\%$} & \textbf{$1\%$} & \textbf{$10\%$}& \textbf{$1\%$} & \textbf{$10\%$}\\ 
\midrule
FedAvg & 1 & ConVIRT & 82.2 & 83.3 & 78.0 & 88.5 & 69.6& 71.5 & GLoRIA & 83.2 & 83.8 & 78.5 & 89.0 & 71.4 & 72.4 \\
FedDRA (ours) & 1 & ConVIRT & 83.2 & 83.7& 81.0 & 90.3 & 71.7 & 74.1 & GLoRIA & 81.8 & 82.9 & 78.5 & 89.6 & 69.3 & 72.5 \\

FedAvg & 5 & ConVIRT & 80.7 & 82.3 & 77.6 & 88.2 & 68.5 & 71.7& GLoRIA & 81.8 & 82.9 & 78.5 & 89.6 & 69.3 & 72.5\\
FedDRA (ours) & 5 & ConVIRT& 81.8 & 82.9 & 78.5 & 89.6 & 69.3 & 72.5 & GLoRIA & 82.2 & 83.0 & 78.4 & 89.4 & 71.5 & 72.9\\
\midrule
Centralized & - & ConVIRT & 83.4 & 84.6 & 82.5 & 92.0 & 72.6 & 76.4 & GLoRIA & 84.0 & 84.7 & 82.2 & 91.8 & 73.5 & 73.7\\
\bottomrule
\end{tabular}
}
\label{tab:hetero main}
\end{table*}

\textbf{FedDRA dynamically schedules updating stepsize for each client, and therefore optimizes the worst-case performance.} We select two clients $C_1, C_2$ in the federated pre-training. For each client, we calculate the average cosine similarity between image and text embeddings, using the server-aggregated model at each communication turn. In Fig.~\ref{fig:all_ana}(d) and Fig.~\ref{fig:all_ana}(e), we plot curves of these similarities, which reflect the cross-modal alignment degree. At each communication turn, when the similarity of a client is relatively high, its similarity in next turn would get a smaller improvement. That's because our FedDRA can assign higher updating stepsize to client where the cross-modal alignment is less extracted by the model. 

\textbf{Our FedDRA can alleviate over-fitting client-specific information, and learn better cross-modal alignment } For our FedDRA and the FedAvg method, we plot the cosine-similarities of text and image embedding averaged across clients at each communication turn. As shown in Fig.\ref{fig:all_ana}(f), FedAvg requires fewer communication rounds to converge, but results in fluctuations after certain communication turns. This aligns with findings in Sec.\ref{empirical finding}, where local retraining a model that are trained after multiple communication rounds, would introduce harmful client-specific information and distort the learned representation space. In contrast, our FedDRA gradually extract cross-modal alignment from local training in a distributionally robust manner, and learns a stronger representation space.

\textbf{Analysis on global constraint hyper-parameter $\mu$.} As shown in Fig.~\ref{fig:all_ana}(c), a larger \( \mu \) encourages federated pre-training to enhance performance on less optimized client data domains, leading to smaller disparity on image-text retrieval performance on each client domain. As we increase $\mu$ from $1$ to $5$, the downstream performance consistently increases. However, a excessively large $\mu$ can decrease overall performance.

\textbf{A larger uncertainty radius $\rho$ improves transferability in downstream tasks.} Fig.~\ref{fig:all_ana}(b) shows the downstream performance of models pre-trained with different uncertainty radii in the DRO process. As larger $\rho$ would bring higher performance in few-shot classification and segmentation tasks on out-of-domain datasets. We also observed that a smaller $\rho$ better supports cross-modal alignment learning, achieving better image-text retrieval performance on in-domain datasets, as shown in Table~\ref{tab:detail rho} in the Appendix. This is because the larger uncertainty radius would incorporate more potential out-of-distribution cases, which can enhance the model's transferability.

\textbf{Robust check on heterogeneity degree of client datasets.} We changed the $\alpha$ which adjusts the heterogeneity degree of the LDA allocated client datasets, to check the robustness of our method under different heterogeneity degree. As shown in Table~\ref{tab:hetero main}, our method consistently enhance pre-training methods' performances under client datasets with different heterogeneity degrees.

\textbf{Robust check on number of clients.} We adjusted the number of clients involved in federated pre-training. As shown in Fig.~\ref{fig:all_ana}(a), increasing the number of clients introduces greater diversity, which can enhance the downstream performance of the pre-trained model.

\section{Conclusion}
\label{conclusion}
Data limitation is a long-standing problem in the multi-modal learning domain. Despite federated learning can leveraging datasets from multiple sources while guaranteeing privacy issues, its performance would be damaged by data heterogeneity. Inspired by our empirical findings on the impact of heterogeneity on federated multi-modal learning, we propose the FedDRA framework to mitigate heterogeneity for federated medical vision-language pre-training. The effectiveness of our method has been verified by comprehensive experiments. While introducing representation from other clients might bring larger improvement, we still consider the most privacy-preserved setting where representations are not transmissible. Further work could explore how to introduce diversity of multi-modal pre-training data while keeping local data private.

\clearpage
%\section*{References}

%{\small
%\bibliographystyle{ieee_fullname}
%\bibliography{main}

\begin{thebibliography}{51}
	\providecommand{\natexlab}[1]{#1}
	\providecommand{\url}[1]{\texttt{#1}}
	\expandafter\ifx\csname urlstyle\endcsname\relax
	\providecommand{\doi}[1]{doi: #1}\else
	\providecommand{\doi}{doi: \begingroup \urlstyle{rm}\Url}\fi
	
	\bibitem[Bannur et~al.(2023)Bannur, Hyland, Liu, Perez-Garcia, Ilse, Castro, Boecking, Sharma, Bouzid, Thieme, et~al.]{bannur2023learning}
	Shruthi Bannur, Stephanie Hyland, Qianchu Liu, Fernando Perez-Garcia, Maximilian Ilse, Daniel~C Castro, Benedikt Boecking, Harshita Sharma, Kenza Bouzid, Anja Thieme, et~al.
	\newblock Learning to exploit temporal structure for biomedical vision-language processing.
	\newblock In \emph{Proceedings of the IEEE/CVF Conference on Computer Vision and Pattern Recognition}, pp.\  15016--15027, 2023.
	
	\bibitem[Bao et~al.(2022)Bao, Wang, Dong, Liu, Mohammed, Aggarwal, Som, Piao, and Wei]{bao2022vlmo}
	Hangbo Bao, Wenhui Wang, Li~Dong, Qiang Liu, Owais~Khan Mohammed, Kriti Aggarwal, Subhojit Som, Songhao Piao, and Furu Wei.
	\newblock Vlmo: Unified vision-language pre-training with mixture-of-modality-experts.
	\newblock \emph{Advances in Neural Information Processing Systems}, 35:\penalty0 32897--32912, 2022.
	
	\bibitem[Bigolin~Lanfredi et~al.(2022)Bigolin~Lanfredi, Zhang, Auffermann, Chan, Duong, Srikumar, Drew, Schroeder, and Tasdizen]{bigolin2022reflacx}
	Ricardo Bigolin~Lanfredi, Mingyuan Zhang, William~F Auffermann, Jessica Chan, Phuong-Anh~T Duong, Vivek Srikumar, Trafton Drew, Joyce~D Schroeder, and Tolga Tasdizen.
	\newblock Reflacx, a dataset of reports and eye-tracking data for localization of abnormalities in chest x-rays.
	\newblock \emph{Scientific data}, 9\penalty0 (1):\penalty0 350, 2022.
	
	\bibitem[Blei et~al.(2003)Blei, Ng, and Jordan]{blei2003latent}
	David~M Blei, Andrew~Y Ng, and Michael~I Jordan.
	\newblock Latent dirichlet allocation.
	\newblock \emph{Journal of machine Learning research}, 3\penalty0 (Jan):\penalty0 993--1022, 2003.
	
	\bibitem[Capitani et~al.(2024)Capitani, Bolelli, Porrello, Calderara, and Ficarra]{capitani2024clusterfix}
	Giacomo Capitani, Federico Bolelli, Angelo Porrello, Simone Calderara, and Elisa Ficarra.
	\newblock Clusterfix: A cluster-based debiasing approach without protected-group supervision.
	\newblock In \emph{Proceedings of the IEEE/CVF Winter Conference on Applications of Computer Vision}, pp.\  4870--4879, 2024.
	
	\bibitem[Castrejon et~al.(2016)Castrejon, Aytar, Vondrick, Pirsiavash, and Torralba]{castrejon2016learning}
	Lluis Castrejon, Yusuf Aytar, Carl Vondrick, Hamed Pirsiavash, and Antonio Torralba.
	\newblock Learning aligned cross-modal representations from weakly aligned data.
	\newblock In \emph{Proceedings of the IEEE conference on computer vision and pattern recognition}, pp.\  2940--2949, 2016.
	
	\bibitem[Chen et~al.(2020)Chen, Gan, Cheng, Li, Carin, and Liu]{chen2020graph}
	Liqun Chen, Zhe Gan, Yu~Cheng, Linjie Li, Lawrence Carin, and Jingjing Liu.
	\newblock Graph optimal transport for cross-domain alignment.
	\newblock In \emph{International Conference on Machine Learning}, pp.\  1542--1553. PMLR, 2020.
	
	\bibitem[Chen et~al.(2022)Chen, Du, Hu, Liu, Li, Wan, and Chang]{chen2022multi}
	Zhihong Chen, Yuhao Du, Jinpeng Hu, Yang Liu, Guanbin Li, Xiang Wan, and Tsung-Hui Chang.
	\newblock Multi-modal masked autoencoders for medical vision-and-language pre-training.
	\newblock In \emph{International Conference on Medical Image Computing and Computer-Assisted Intervention}, pp.\  679--689. Springer, 2022.
	
	\bibitem[Collins et~al.(2021)Collins, Hassani, Mokhtari, and Shakkottai]{collins2021exploiting}
	Liam Collins, Hamed Hassani, Aryan Mokhtari, and Sanjay Shakkottai.
	\newblock Exploiting shared representations for personalized federated learning.
	\newblock In \emph{International conference on machine learning}, pp.\  2089--2099. PMLR, 2021.
	
	\bibitem[Decenci{\`e}re et~al.(2014)Decenci{\`e}re, Zhang, Cazuguel, Lay, Cochener, Trone, Gain, Ord{\'o}{\~n}ez-Varela, Massin, Erginay, et~al.]{decenciere2014feedback}
	Etienne Decenci{\`e}re, Xiwei Zhang, Guy Cazuguel, Bruno Lay, B{\'e}atrice Cochener, Caroline Trone, Philippe Gain, John-Richard Ord{\'o}{\~n}ez-Varela, Pascale Massin, Ali Erginay, et~al.
	\newblock Feedback on a publicly distributed image database: the messidor database.
	\newblock \emph{Image Analysis \& Stereology}, pp.\  231--234, 2014.
	
	\bibitem[Deng et~al.(2020)Deng, Kamani, and Mahdavi]{deng2020distributionally}
	Yuyang Deng, Mohammad~Mahdi Kamani, and Mehrdad Mahdavi.
	\newblock Distributionally robust federated averaging.
	\newblock \emph{Advances in neural information processing systems}, 33:\penalty0 15111--15122, 2020.
	
	\bibitem[Devlin et~al.(2018)Devlin, Chang, Lee, and Toutanova]{devlin2018bert}
	Jacob Devlin, Ming-Wei Chang, Kenton Lee, and Kristina Toutanova.
	\newblock Bert: Pre-training of deep bidirectional transformers for language understanding.
	\newblock \emph{arXiv preprint arXiv:1810.04805}, 2018.
	
	\bibitem[Dosovitskiy et~al.(2020)Dosovitskiy, Beyer, Kolesnikov, Weissenborn, Zhai, Unterthiner, Dehghani, Minderer, Heigold, Gelly, et~al.]{dosovitskiy2020image}
	Alexey Dosovitskiy, Lucas Beyer, Alexander Kolesnikov, Dirk Weissenborn, Xiaohua Zhai, Thomas Unterthiner, Mostafa Dehghani, Matthias Minderer, Georg Heigold, Sylvain Gelly, et~al.
	\newblock An image is worth 16x16 words: Transformers for image recognition at scale.
	\newblock \emph{arXiv preprint arXiv:2010.11929}, 2020.
	
	\bibitem[Gao et~al.(2024)Gao, Liu, Xu, Wu, Zhang, Li, Yang, Liu, and Sun]{gao2024softclip}
	Yuting Gao, Jinfeng Liu, Zihan Xu, Tong Wu, Enwei Zhang, Ke~Li, Jie Yang, Wei Liu, and Xing Sun.
	\newblock Softclip: Softer cross-modal alignment makes clip stronger.
	\newblock In \emph{Proceedings of the AAAI Conference on Artificial Intelligence}, volume~38, pp.\  1860--1868, 2024.
	
	\bibitem[Ghosh et~al.(2019)Ghosh, Hong, Yin, and Ramchandran]{ghosh2019robust}
	Avishek Ghosh, Justin Hong, Dong Yin, and Kannan Ramchandran.
	\newblock Robust federated learning in a heterogeneous environment.
	\newblock \emph{arXiv preprint arXiv:1906.06629}, 2019.
	
	\bibitem[Han et~al.(2023)Han, Liu, Liu, and Xiong]{han2023distributionally}
	Peixuan Han, Zhenghao Liu, Zhiyuan Liu, and Chenyan Xiong.
	\newblock Distributionally robust unsupervised dense retrieval training on web graphs.
	\newblock \emph{arXiv preprint arXiv:2310.16605}, 2023.
	
	\bibitem[Han et~al.(2022)Han, Park, Wu, Kim, Wu, Xie, and Cha]{han2022fedx}
	Sungwon Han, Sungwon Park, Fangzhao Wu, Sundong Kim, Chuhan Wu, Xing Xie, and Meeyoung Cha.
	\newblock Fedx: Unsupervised federated learning with cross knowledge distillation.
	\newblock In \emph{European Conference on Computer Vision}, pp.\  691--707. Springer, 2022.
	
	\bibitem[Huang et~al.(2021)Huang, Shen, Lungren, and Yeung]{huang2021gloria}
	Shih-Cheng Huang, Liyue Shen, Matthew~P Lungren, and Serena Yeung.
	\newblock Gloria: A multimodal global-local representation learning framework for label-efficient medical image recognition.
	\newblock In \emph{Proceedings of the IEEE/CVF International Conference on Computer Vision}, pp.\  3942--3951, 2021.
	
	\bibitem[Huang et~al.(2022)Huang, Ye, and Du]{huang2022learn}
	Wenke Huang, Mang Ye, and Bo~Du.
	\newblock Learn from others and be yourself in heterogeneous federated learning.
	\newblock In \emph{Proceedings of the IEEE/CVF Conference on Computer Vision and Pattern Recognition}, pp.\  10143--10153, 2022.
	
	\bibitem[Ladbury et~al.(2023)Ladbury, Amini, Govindarajan, Mambetsariev, Raz, Massarelli, Williams, Rodin, and Salgia]{ladbury2023integration}
	Colton Ladbury, Arya Amini, Ameish Govindarajan, Isa Mambetsariev, Dan~J Raz, Erminia Massarelli, Terence Williams, Andrei Rodin, and Ravi Salgia.
	\newblock Integration of artificial intelligence in lung cancer: Rise of the machine.
	\newblock \emph{Cell Reports Medicine}, 2023.
	
	\bibitem[Legate et~al.(2024)Legate, Bernier, Page-Caccia, Oyallon, and Belilovsky]{legate2024guiding}
	Gwen Legate, Nicolas Bernier, Lucas Page-Caccia, Edouard Oyallon, and Eugene Belilovsky.
	\newblock Guiding the last layer in federated learning with pre-trained models.
	\newblock \emph{Advances in Neural Information Processing Systems}, 36, 2024.
	
	\bibitem[Levy et~al.(2020)Levy, Carmon, Duchi, and Sidford]{levy2020large}
	Daniel Levy, Yair Carmon, John~C Duchi, and Aaron Sidford.
	\newblock Large-scale methods for distributionally robust optimization.
	\newblock \emph{Advances in Neural Information Processing Systems}, 33:\penalty0 8847--8860, 2020.
	
	\bibitem[Li \& Wang(2019)Li and Wang]{li2019fedmd}
	Daliang Li and Junpu Wang.
	\newblock Fedmd: Heterogenous federated learning via model distillation.
	\newblock \emph{arXiv preprint arXiv:1910.03581}, 2019.
	
	\bibitem[Li et~al.(2022{\natexlab{a}})Li, Lyu, Iso, Chakrabarti, and Spranger]{li2022mocosfl}
	Jingtao Li, Lingjuan Lyu, Daisuke Iso, Chaitali Chakrabarti, and Michael Spranger.
	\newblock Mocosfl: enabling cross-client collaborative self-supervised learning.
	\newblock In \emph{The Eleventh International Conference on Learning Representations}, 2022{\natexlab{a}}.
	
	\bibitem[Li et~al.(2021{\natexlab{a}})Li, Selvaraju, Gotmare, Joty, Xiong, and Hoi]{li2021align}
	Junnan Li, Ramprasaath Selvaraju, Akhilesh Gotmare, Shafiq Joty, Caiming Xiong, and Steven Chu~Hong Hoi.
	\newblock Align before fuse: Vision and language representation learning with momentum distillation.
	\newblock \emph{Advances in neural information processing systems}, 34:\penalty0 9694--9705, 2021{\natexlab{a}}.
	
	\bibitem[Li et~al.(2022{\natexlab{b}})Li, Li, Xiong, and Hoi]{li2022blip}
	Junnan Li, Dongxu Li, Caiming Xiong, and Steven Hoi.
	\newblock Blip: Bootstrapping language-image pre-training for unified vision-language understanding and generation.
	\newblock In \emph{International Conference on Machine Learning}, pp.\  12888--12900. PMLR, 2022{\natexlab{b}}.
	
	\bibitem[Li et~al.(2021{\natexlab{b}})Li, He, and Song]{li2021model}
	Qinbin Li, Bingsheng He, and Dawn Song.
	\newblock Model-contrastive federated learning.
	\newblock In \emph{Proceedings of the IEEE/CVF conference on computer vision and pattern recognition}, pp.\  10713--10722, 2021{\natexlab{b}}.
	
	\bibitem[Liu et~al.(2021)Liu, Fu, Xu, Yang, Li, Wang, and Zhang]{liu2021learning}
	Chen Liu, Yanwei Fu, Chengming Xu, Siqian Yang, Jilin Li, Chengjie Wang, and Li~Zhang.
	\newblock Learning a few-shot embedding model with contrastive learning.
	\newblock In \emph{Proceedings of the AAAI conference on artificial intelligence}, volume~35, pp.\  8635--8643, 2021.
	
	\bibitem[Liu et~al.(2022)Liu, Shen, Cui, Zhou, Kuang, and Li]{liu2022distributionally}
	Jiashuo Liu, Zheyan Shen, Peng Cui, Linjun Zhou, Kun Kuang, and Bo~Li.
	\newblock Distributionally robust learning with stable adversarial training.
	\newblock \emph{IEEE Transactions on Knowledge and Data Engineering}, 2022.
	
	\bibitem[Lu et~al.(2023)Lu, Liu, Liu, and Zhou]{lu2023scaling}
	Siyu Lu, Zheng Liu, Tianlin Liu, and Wangchunshu Zhou.
	\newblock Scaling-up medical vision-and-language representation learning with federated learning.
	\newblock \emph{Engineering Applications of Artificial Intelligence}, 126:\penalty0 107037, 2023.
	
	\bibitem[Lu et~al.(2024)Lu, Zhang, Sun, Guo, and Yu]{lu2024f}
	Yiwei Lu, Guojun Zhang, Sun Sun, Hongyu Guo, and Yaoliang Yu.
	\newblock $ f $-micl: Understanding and generalizing infonce-based contrastive learning.
	\newblock \emph{arXiv preprint arXiv:2402.10150}, 2024.
	
	\bibitem[McMahan et~al.(2017)McMahan, Moore, Ramage, Hampson, and y~Arcas]{mcmahan2017communication}
	Brendan McMahan, Eider Moore, Daniel Ramage, Seth Hampson, and Blaise~Aguera y~Arcas.
	\newblock Communication-efficient learning of deep networks from decentralized data.
	\newblock In \emph{Artificial intelligence and statistics}, pp.\  1273--1282. PMLR, 2017.
	
	\bibitem[Mendogni et~al.(2020)Mendogni, Vannucci, Ghisalberti, Anile, Aramini, Congedo, Nosotti, Bertolaccini, Collaborators of~the Pneumothorax Working~Group, D’Ambrosio, et~al.]{mendogni2020epidemiology}
	Paolo Mendogni, Jacopo Vannucci, Marco Ghisalberti, Marco Anile, Beatrice Aramini, Maria~Teresa Congedo, Mario Nosotti, Luca Bertolaccini, on~behalf of the Italian Society for Thoracic Surgery (endorsed by the Italian Ministry of Health) Collaborators of the Pneumothorax Working~Group Collaborators of~the Pneumothorax Working~Group, Ambra~Enrica D’Ambrosio, et~al.
	\newblock Epidemiology and management of primary spontaneous pneumothorax: a systematic review.
	\newblock \emph{Interactive cardiovascular and thoracic surgery}, 30\penalty0 (3):\penalty0 337--345, 2020.
	
	\bibitem[Nakayama et~al.(2023)Nakayama, Goncalves, Zago~Ribeiro, Santos, Ferraz, Malerbi, Celi, and Regatieri]{nakayama2023brazilian}
	Luis~Filipe Nakayama, Mariana Goncalves, L~Zago~Ribeiro, Helen Santos, Daniel Ferraz, Fernando Malerbi, Leo~Anthony Celi, and Caio Regatieri.
	\newblock A brazilian multilabel ophthalmological dataset (brset).
	\newblock \emph{PhysioNet https://doi. org/10}, 13026, 2023.
	
	\bibitem[Nakayama et~al.()]{nakayamambrset}
	Luis~Filipe Nakayama et~al.
	\newblock mbrset, a mobile brazilian retinal dataset.
	
	\bibitem[Oquab et~al.(2023)Oquab, Darcet, Moutakanni, Vo, Szafraniec, Khalidov, Fernandez, Haziza, Massa, El-Nouby, et~al.]{oquab2023dinov2}
	Maxime Oquab, Timoth{\'e}e Darcet, Th{\'e}o Moutakanni, Huy Vo, Marc Szafraniec, Vasil Khalidov, Pierre Fernandez, Daniel Haziza, Francisco Massa, Alaaeldin El-Nouby, et~al.
	\newblock Dinov2: Learning robust visual features without supervision.
	\newblock \emph{arXiv preprint arXiv:2304.07193}, 2023.
	
	\bibitem[Radford et~al.(2021)Radford, Kim, Hallacy, Ramesh, Goh, Agarwal, Sastry, Askell, Mishkin, Clark, et~al.]{radford2021learning}
	Alec Radford, Jong~Wook Kim, Chris Hallacy, Aditya Ramesh, Gabriel Goh, Sandhini Agarwal, Girish Sastry, Amanda Askell, Pamela Mishkin, Jack Clark, et~al.
	\newblock Learning transferable visual models from natural language supervision.
	\newblock In \emph{International conference on machine learning}, pp.\  8748--8763. PMLR, 2021.
	
	\bibitem[Rahimian \& Mehrotra(2019)Rahimian and Mehrotra]{rahimian2019distributionally}
	Hamed Rahimian and Sanjay Mehrotra.
	\newblock Distributionally robust optimization: A review.
	\newblock \emph{arXiv preprint arXiv:1908.05659}, 2019.
	
	\bibitem[Rehman et~al.(2023)Rehman, Gao, De~Gusm{\~a}o, Alibeigi, Shen, and Lane]{rehman2023dawa}
	Yasar Abbas~Ur Rehman, Yan Gao, Pedro Porto~Buarque De~Gusm{\~a}o, Mina Alibeigi, Jiajun Shen, and Nicholas~D Lane.
	\newblock L-dawa: Layer-wise divergence aware weight aggregation in federated self-supervised visual representation learning.
	\newblock In \emph{Proceedings of the IEEE/CVF international conference on computer vision}, pp.\  16464--16473, 2023.
	
	\bibitem[Saab et~al.(2022)Saab, Hooper, Chen, Zhang, Rubin, and R{\'e}]{saab2022reducing}
	Khaled Saab, Sarah Hooper, Mayee Chen, Michael Zhang, Daniel Rubin, and Christopher R{\'e}.
	\newblock Reducing reliance on spurious features in medical image classification with spatial specificity.
	\newblock In \emph{Machine Learning for Healthcare Conference}, pp.\  760--784. PMLR, 2022.
	
	\bibitem[Shih et~al.(2019)Shih, Wu, Halabi, Kohli, Prevedello, Cook, Sharma, Amorosa, Arteaga, Galperin-Aizenberg, et~al.]{shih2019augmenting}
	George Shih, Carol~C Wu, Safwan~S Halabi, Marc~D Kohli, Luciano~M Prevedello, Tessa~S Cook, Arjun Sharma, Judith~K Amorosa, Veronica Arteaga, Maya Galperin-Aizenberg, et~al.
	\newblock Augmenting the national institutes of health chest radiograph dataset with expert annotations of possible pneumonia.
	\newblock \emph{Radiology: Artificial Intelligence}, 1\penalty0 (1):\penalty0 e180041, 2019.
	
	\bibitem[Su et~al.(2023)Su, Zhu, Tao, Lu, Li, Huang, Qiao, Wang, Zhou, and Dai]{su2023towards}
	Weijie Su, Xizhou Zhu, Chenxin Tao, Lewei Lu, Bin Li, Gao Huang, Yu~Qiao, Xiaogang Wang, Jie Zhou, and Jifeng Dai.
	\newblock Towards all-in-one pre-training via maximizing multi-modal mutual information.
	\newblock In \emph{Proceedings of the IEEE/CVF Conference on Computer Vision and Pattern Recognition}, pp.\  15888--15899, 2023.
	
	\bibitem[Wang et~al.(2022)Wang, Zhou, Wang, Vardhanabhuti, and Yu]{wang2022multi}
	Fuying Wang, Yuyin Zhou, Shujun Wang, Varut Vardhanabhuti, and Lequan Yu.
	\newblock Multi-granularity cross-modal alignment for generalized medical visual representation learning.
	\newblock \emph{Advances in Neural Information Processing Systems}, 35:\penalty0 33536--33549, 2022.
	
	\bibitem[Wang et~al.(2020)Wang, Lin, and Wong]{wang2020covid}
	Linda Wang, Zhong~Qiu Lin, and Alexander Wong.
	\newblock Covid-net: A tailored deep convolutional neural network design for detection of covid-19 cases from chest x-ray images.
	\newblock \emph{Scientific reports}, 10\penalty0 (1):\penalty0 19549, 2020.
	
	\bibitem[Yan et~al.(2023)Yan, Qu, Wei, Huang, Shen, Rubin, Xing, and Zhou]{yan2023label}
	Rui Yan, Liangqiong Qu, Qingyue Wei, Shih-Cheng Huang, Liyue Shen, Daniel Rubin, Lei Xing, and Yuyin Zhou.
	\newblock Label-efficient self-supervised federated learning for tackling data heterogeneity in medical imaging.
	\newblock \emph{IEEE Transactions on Medical Imaging}, 2023.
	
	\bibitem[Zhang et~al.(2023)Zhang, Kuang, Chen, You, Shen, Xiao, Zhang, Wu, Wu, Zhuang, et~al.]{zhang2023federated}
	Fengda Zhang, Kun Kuang, Long Chen, Zhaoyang You, Tao Shen, Jun Xiao, Yin Zhang, Chao Wu, Fei Wu, Yueting Zhuang, et~al.
	\newblock Federated unsupervised representation learning.
	\newblock \emph{Frontiers of Information Technology \& Electronic Engineering}, 24\penalty0 (8):\penalty0 1181--1193, 2023.
	
	\bibitem[Zhang et~al.(2024{\natexlab{a}})Zhang, Hua, Cao, Wang, Song, Xue, Ma, and Guan]{zhang2024eliminating}
	Jianqing Zhang, Yang Hua, Jian Cao, Hao Wang, Tao Song, Zhengui Xue, Ruhui Ma, and Haibing Guan.
	\newblock Eliminating domain bias for federated learning in representation space.
	\newblock \emph{Advances in Neural Information Processing Systems}, 36, 2024{\natexlab{a}}.
	
	\bibitem[Zhang et~al.(2024{\natexlab{b}})Zhang, Xu, Chen, Xie, and Chen]{zhang2024prototypical}
	Yilan Zhang, Yingxue Xu, Jianqi Chen, Fengying Xie, and Hao Chen.
	\newblock Prototypical information bottlenecking and disentangling for multimodal cancer survival prediction.
	\newblock \emph{arXiv preprint arXiv:2401.01646}, 2024{\natexlab{b}}.
	
	\bibitem[Zhang et~al.(2022)Zhang, Jiang, Miura, Manning, and Langlotz]{zhang2022contrastive}
	Yuhao Zhang, Hang Jiang, Yasuhide Miura, Christopher~D Manning, and Curtis~P Langlotz.
	\newblock Contrastive learning of medical visual representations from paired images and text.
	\newblock In \emph{Machine Learning for Healthcare Conference}, pp.\  2--25. PMLR, 2022.
	
	\bibitem[Zhuang et~al.(2021)Zhuang, Gan, Wen, Zhang, and Yi]{zhuang2021collaborative}
	Weiming Zhuang, Xin Gan, Yonggang Wen, Shuai Zhang, and Shuai Yi.
	\newblock Collaborative unsupervised visual representation learning from decentralized data.
	\newblock In \emph{Proceedings of the IEEE/CVF international conference on computer vision}, pp.\  4912--4921, 2021.
	
	\bibitem[Zhuang et~al.(2022)Zhuang, Wen, and Zhang]{zhuang2022divergence}
	Weiming Zhuang, Yonggang Wen, and Shuai Zhang.
	\newblock Divergence-aware federated self-supervised learning.
	\newblock \emph{arXiv preprint arXiv:2204.04385}, 2022.
	
\end{thebibliography}
%}
%%%%%%%%%%%%%%%%%%%%%%%%%%%%%%%%%%%%%%%%%%%%%%%%%%%%%%%%%%%%
\clearpage
\appendix

\section{Implementation Detailed}
\subsection{Details of MIMIC-CXR}
\subsubsection{Pre-training setup}
\label{detailed setup}

Following~\citep{wang2022multi} we utilize the MIMIC-CXR \citep{bigolin2022reflacx} dataset for multi-modal pre-training. This dataset is widely used in the medical multi-modal learning domain, with $227,835$ image-text pairs from $65,379$ patients. Some related works also have imported additional features to image-text pairs to augment the data. However, we only use the image-text pairs for pre-training to make the results and conclusions more generalizable. The MIMIC-CXR dataset is open access, it can be obtained through \href{https://physionet.org/content/mimic-cxr/2.0.0/}{MIMIC-CXR Access}.

\begin{figure*}[h] 
    \centering
    \includegraphics[width=\textwidth]{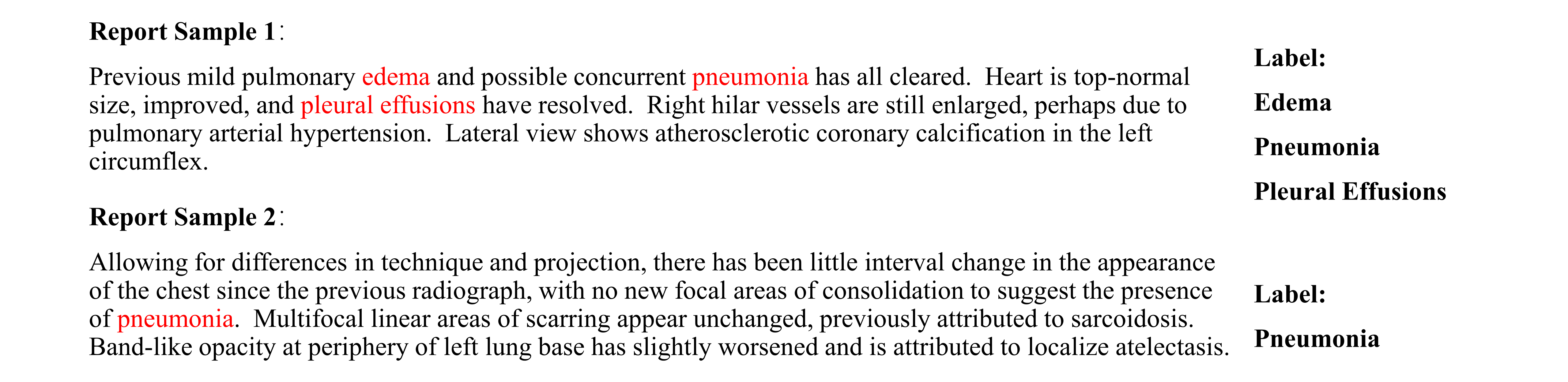} % Adjust the path and width as needed

    \caption{Illustration of the strong connection between latent variable and the text modality.}

    \label{fig:text data}
\end{figure*}

During the pre-training, local clients only have access to their highly heterogeneous datasets. To construct the heterogeneous client datasets, following~\citep{yan2023label} we employ the Latent Dirichlet Allocation (LDA)~\citep{blei2003latent} to divide the MIMIC-CXR dataset into $5$ partitions based on a selected sensitive attribute. For implementation, we import the corresponding attribute information of given image-text pairs from the MIMIC-CXR and divide local datasets based on disease category. The disease category is a multi-label binary attribute and is transformed into a multi-class label. That's because the words in the clinical report are highly related to the disease category as illustrated in Fig~\ref{fig:text data}. We set the heterogeneity degree in the LDA algorithm to be 1 for main experiments. For analysis experiments, we also have run experiments on client datasets allocated by LDA with a heterogeneity degree of 5.

Specifically, we select 5 commonly considered diseases~\cite{bannur2023learning}: 'Edema','Pleural Effusion', 'Consolidation', 'Pneumothorax', and 'Pneumonia'. We set the non-NaN value to 1 and then set NaN value to 0 to construct a 5-way  binary multi-label. Then we get $2^5$-category multi-class label and run LDA on them. %The distribution of the disease of the client dataset is summarized in Table~\ref{subgroup}.

We divide the MIMIC-CXR into 5 heterogeneous subgroups to construct 5 client datasets. Each divided dataset consists of train splits and test splits based on the notation of the MIMIC-CXR. Our pre-trainings are mainly conducted on $4\times A40$ or $2\times A100$. The batch size we have utilized ranged from $288$ to 388. We set the learning rate to $2\times 10^{-5}$ in main experiments, the number of communications to 25. For our method, we set the uncertainty radius $\rho=0.1,\mu =5$ in main experiments. For each communication, we randomly sample 50 batches of data from the client datasets.

\subsubsection{Downstream tasks} \label{detailed downstream} We evaluate the generalization ability of the pre-trained model through three downstream tasks: few-shot classification, medical image segmentation, and image retrieval.

\textbf{Few-shot classification.} To assess the model's effectiveness on general medical image tasks, we evaluate it on multiple image classification benchmarks: (1) RSNA Pneumonia Detection (RSNA)\cite{shih2019augmenting}, where the task is to predict whether an image shows pneumonia. (2) Covidx\cite{wang2020covid}, which includes three categories: COVID-19, non-COVID pneumonia, and normal. We fine-tune our pre-trained model with an additional linear layer on $1\%$ and $10\%$ of the training dataset and report classification accuracy on these benchmarks.

\textbf{Medical image segmentation.} To explore the model’s transferability to fine-grained tasks, we conduct experiments on medical image segmentation using the RSNA~\cite{wang2020covid} benchmark. Following~\cite{wang2022multi}, we convert RSNA object detection ground truths into segmentation masks. Similar to~\cite{huang2021gloria}, we employ a U-Net framework with our pre-trained image encoder as the frozen encoder, while fine-tuning the decoder on $1\%$ and $10\%$ of the training data. The Dice score is used for performance evaluation.

\textbf{Image retrieval.} To verify whether the pre-trained models have captured the semantic alignment between image and text in the pre-training data, we perform an image retrieval task. We test image retrieval performance on the validation splits of the local clients. For each text in a batch of image-text pairs, we calculate similarities with images in the batch, then rank these similarities and retrieve the top-1 and top-5 images. If the corresponding image of the text is in the selected set, it is correctly retrieved. We use top-1 and top-5 recall accuracy to evaluate performance.

\subsection{Ophthalmology datasets}
\subsubsection{Pre-training setup.} 
We conduct vision-language multi-modal pre-training using retinal image datasets from different institutes. These retinal datasets are from different institutions of low-income and high-income countries, and are highly heterogeneous real-world scenes. Specifically, we utilize MESSIDOR~\citep{decenciere2014feedback} from France and BRSET~\citep{nakayama2023brazilian} from Brazil as pre-training datasets, and assign them to two clients. These datasets include tabular EHR records indicating Diabetic Retinopathy (DR) status and edema risk. We transform tabular data into text captions in the format: "retinal image with $\{$DR status$\}$ and $\{$edema risk$\}$" to obtain text prompts. Similar to MIMIC dataset, our pre-trainings on ophthalmology datasets are mainly conducted on $4\times A40$ or $2\times A100$. We set the batch size to 100, the number of communications to 20, and the learning rate to $1\times 10^{-5}$ in the experiments. For our method, we set the uncertainty radius $\rho=0.5,\mu =1$ in main experiments. For each communication, we randomly sample 20 batches of data from the client datasets. 

\subsubsection{Downstream tasks.} 
We evaluate the transferability of the models on few-shot classification tasks using the MBRSET~\citep{nakayamambrset} dataset. Unlike the pre-training datasets, MBRSET was collected in low-income areas using portable devices, resulting in a significant distribution shift. We perform few-shot classification tasks on diabetic retinopathy and edema status prediction tasks using this dataset. These are binary classification problems. We fine-tune the model with an additional linear layer on $10\%$, $20\%$ and $100\%$ of the training data, and report classification accuracies.
\clearpage

\section{Additional Experiment Results}
\textbf{Federated pre-trained models still show a significant performance gap compared to centralized pre-trained models in multi-modal retrieval tasks.} Table~\ref{tab:multimethod_main} shows the performance of models pre-trained in decentralized, FedAvg, centralized federated learning strategies, using different backbone pre-training methods. FedAvg has more effectively extract cross-modal alignment from federally utilizing local datasets, and achieved much better transferability on downstream datasets and in-domain image-text retrieval tasks, compared to de-centralized pre-trained models. However, there are still performance gaps in the retrieval tasks compared to the centralized pre-trained model. That might because each batch of data in centralized pre-training scene has higher diversity, which encourages the contrastive-based model to capture more robust alignment.  

\begin{table*}[h] % The 't' means "top of the page"
\centering % This centers the table

\caption{Downstream task performance on different multi-modal pre-training backbone methods.}
\resizebox{\textwidth}{!}{
\begin{tabular}{@{} ll ll ll  ll  llll }
\toprule
\textbf{Strategy} & \textbf{Backbone} &\multicolumn{2}{c}{\textbf{RSNA (cls.)}} &\multicolumn{2}{c}{\textbf{Covid (cls.)}} &\multicolumn{2}{c}{\textbf{RSNA (seg.)}}  &\multicolumn{4}{c}{\textbf{In-domain Image-Text Retrieval}}\\
\cmidrule(l){3-4}  \cmidrule(l){5-6}  \cmidrule(l){7-8}  \cmidrule(l){9-12}  
 & & \textbf{$1\%$} & \textbf{$10\%$} & \textbf{$1\%$} & \textbf{$10\%$}  & \textbf{$1\%$} & \textbf{$10\%$}  & \textbf{Rec.@1} & \textbf{Rec.@5} & \textbf{Wst.@1} & \textbf{Wst.@5}\\ 
\midrule
Decentralized & ConVIRT & 81.5 & 82.3 & 76.5 & 85.6 & 64.6 & 70.7 & 15.5 & 51.1& 13.6 & 46.0\\
FedAvg& ConVIRT & 83.1 & 83.3 & 78.0 & 88.5 & 69.6 & 71.5 & 28.8 & 72.1 & 25.3 & 66.7\\
\rowcolor{gray!30} Centralized & ConVIRT & 83.4 & 84.6 & 82.5 & 92.0 & 72.6 & 76.4 & 41.5 & 84.2 & 38.6 & 80.0 \\
\midrule
Decentralized & GLoRIA & 82.3 & 82.9 & 77.9 & 86.8 & 71.1 & 72.1 & 17.2 & 52.5 & 15.2 & 48.7 \\
FedAvg & GLoRIA & 83.2 & 83.3 & 77.5 & 89.0 & 71.4 & 72.4 & 29.9 & 73.8 & 27.8 & 69.5 \\
\midrule
\rowcolor{gray!30}
Centralized & GLoRIA & 84.0 & 84.7 & 82.2 & 91.8 & 73.6 & 73.7 & 41.7 & 84.0 & 39.0 & 80.5\\
\midrule
Decentralized & MGCA& 81.9 & 82.7 & 77.8 & 87.6 & 62.8 & 70.2 & 15.2 & 50.4 & 13.4 & 45.4 \\
FedAvg & MGCA & 82.6 & 83.5 & 75.8 & 88.2 & 70.1 & 71.4 & 29.3 & 73.7 & 26.8 & 70.4\\
\midrule
\rowcolor{gray!30}
Centralized & MGCA & 84.0 & 84.5 & 79.5 & 89.5 & 70.7 & 72.5 & 39.9 & 83.5 & 36.9 & 80.3\\
\bottomrule
\end{tabular}
}
\label{tab:multimethod_main}
\end{table*}

\section{Detailed Experiment Results}
Here we provide detailed results of ablation studies shown in Fig.~\ref{fig:all_ana} in the main text.

\begin{table*}[h] % The 't' means "top of the page"
\centering % This centers the table

\caption{Ablation studies on the number of clients.}
\resizebox{\textwidth}{!}{
\begin{tabular}{@{} l ll ll  ll  llll }
\toprule
\textbf{Num. of Client} &\multicolumn{2}{c}{\textbf{RSNA (cls.)}} &\multicolumn{2}{c}{\textbf{Covid (cls.)}} &\multicolumn{2}{c}{\textbf{RSNA (seg.)}}  &\multicolumn{4}{c}{\textbf{In-domain Image-Text Retrieval}}\\
\cmidrule(l){2-3}  \cmidrule(l){4-5}  \cmidrule(l){6-7}  \cmidrule(l){8-11}  
 & \textbf{$1\%$} & \textbf{$10\%$} & \textbf{$1\%$} & \textbf{$10\%$}  & \textbf{$1\%$} & \textbf{$10\%$}  & \textbf{Rec.@1} & \textbf{Rec.@5} & \textbf{Wst.@1} & \textbf{Wst.@5}\\ 
\midrule
n=2 & 82.1 & 83.2 & 78.4 & 88.5 & 61.8 & 71.0 & 23.1 & 62.9 & 19.2 & 57.8\\
n=5 & \textbf{83.2} & \textbf{83.7} & \textbf{81.0} & \textbf{90.3} & \textbf{71.7} & \textbf{74.1} & \textbf{30.2} & \textbf{73.2} & \textbf{27.0} & \textbf{68.9}\\
\bottomrule
\end{tabular}
}
\label{tab:detail num of client}
\end{table*}

\begin{table*}[h] % The 't' means "top of the page"
\centering % This centers the table

\caption{Ablation studies on uncertainty radius.}
\resizebox{\textwidth}{!}{
\begin{tabular}{@{} l ll ll  ll  llll }
\toprule
\textbf{Uncertainty Radius} &\multicolumn{2}{c}{\textbf{RSNA (cls.)}} &\multicolumn{2}{c}{\textbf{Covid (cls.)}} &\multicolumn{2}{c}{\textbf{RSNA (seg.)}}  &\multicolumn{4}{c}{\textbf{In-domain Image-Text Retrieval}}\\
\cmidrule(l){2-3}  \cmidrule(l){4-5}  \cmidrule(l){6-7}  \cmidrule(l){8-11}  
 & \textbf{$1\%$} & \textbf{$10\%$} & \textbf{$1\%$} & \textbf{$10\%$}  & \textbf{$1\%$} & \textbf{$10\%$}  & \textbf{Rec.@1} & \textbf{Rec.@5} & \textbf{Wst.@1} & \textbf{Wst.@5}\\ 
\midrule
$\rho=$0.01 & 82.7 & 83.2 & 79.6 & 89.1 & 71.0 & 72.8 & \textbf{30.4} & \textbf{73.5} & 26.6 & 68.4\\
$\rho=$0.1 & 83.2 & 83.7 & 81.0 & 90.3 & 71.7 & \textbf{74.1} & 30.2 & 73.2 & \textbf{27.0} & \textbf{68.9}\\
$\rho=$1 & \textbf{83.3} & \textbf{84.0} & \textbf{81.3} & \textbf{90.8} & \textbf{72.1} & 74.1 & 28.9 & 72.5 & 26.2 & 67.8\\
\bottomrule
\end{tabular}
}
\label{tab:detail rho}
\end{table*}

\begin{table*}[h] % The 't' means "top of the page"
\centering % This centers the table

\caption{Ablation studies on global constraint degree.}
\resizebox{\textwidth}{!}{
\begin{tabular}{@{} l ll ll  ll  lll }
\toprule
\textbf{Constraint Degree} &\multicolumn{2}{c}{\textbf{RSNA (cls.)}} &\multicolumn{2}{c}{\textbf{Covid (cls.)}} &\multicolumn{2}{c}{\textbf{RSNA (seg.)}}  &\multicolumn{3}{c}{\textbf{In-domain Image-Text Retrieval}}\\
\cmidrule(l){2-3}  \cmidrule(l){4-5}  \cmidrule(l){6-7}  \cmidrule(l){8-10}  
 & \textbf{$1\%$} & \textbf{$10\%$} & \textbf{$1\%$} & \textbf{$10\%$}  & \textbf{$1\%$} & \textbf{$10\%$}  & \textbf{Rec.@1} & \textbf{Rec.@5} & \textbf{Disparity}\\ 
\midrule
$\mu=$1 & 82.8 & 83.4 & 79.8 & 89.6 & 70.5 & 72.8 & 29.1 & 72.5 & 3.2\\
$\mu=$5 & \textbf{83.2} & \textbf{83.7} & \textbf{81.0} & \textbf{90.3} & \textbf{71.7} & \textbf{74.1} & \textbf{30.2} & \textbf{73.2} & 2.9\\
$\mu=$10 & 82.6 & 83.2 & 80.2 & 90.2 & 71.3 & 72.9 & 29.6 & 72.8 & \textbf{2.4}\\
\bottomrule
\end{tabular}
}
\label{tab:detail mu}
\end{table*}

Here we provided detailed results for our empirical study in Sec.~\ref{empirical finding}.

\begin{table*}[ht!]
\centering
\label{tab:image_retrieval}
\caption{The comparison of retrieval acc. on each client denoted as $\{C_i\}_{i=1}^5$, of centralized, FedAvg, and averaged acc. of decentralized pre-trained models using the ConVIRT backbone. We report the local models of decentralized pre-training strategy as Decentralized$_i$.}
\resizebox{0.95\textwidth}{!}{
\begin{tabular}{@{} llllll lllllll}
\toprule
 \multicolumn{1}{c}{\textbf{Strategy}} & \multicolumn{5}{c}{\textbf{Recall@1 (ACC)}} & \multicolumn{5}{c}{\textbf{Recall@5 (ACC)}}\\
\midrule 
  & \textbf{C1} & \textbf{C2} & \textbf{C3} & \textbf{C4} & \textbf{C5} & \textbf{Avg.} & \textbf{C1} & \textbf{C2} & \textbf{C3} & \textbf{C4} & \textbf{C5} & \textbf{Avg.}\\ 
\midrule
Centralized & 43.6 & 38.6 & 40.1 & 43.1 & 41.9 & 44.4 & 86.6 & 80.0 & 82.6 & 85.0 & 86.8 & 84.2\\
\midrule
FedAvg & 30.4 & 25.3 & 26.9 & 28.8 & 32.7 & 28.8 & 76.4 & 66.7 & 69.8 & 73.8 & 73.9 & 72.1\\
\midrule
Decentralized$_1$ & 17.7 & 14.7 & 15.4 & 18.2 & 14.1 & 16.0 & 57.0 & 49.6 & 51.3 & 54.6 & 55.1 & 53.5\\
Decentralized$_2$ & 15.3 & 11.6 & 13.9 & 15.0 & 13.8 & 13.9 & 54.8 & 41.9 & 46.2 & 47.9 & 45.6 & 47.3\\
Decentralized$_3$ & 17.4 & 13.1 & 14.1 & 15.1 & 15.4 & 15.0 & 50.4 & 44.2 & 46.4 & 49.7 & 52.5 & 48.6\\
Decentralized$_4$ & 16.5 & 14.3 & 14.1 & 15.4 & 17.4 & 15.5 & 57.0 & 45.7 & 47.5 & 52.4 & 57.6 & 52.0\\
Decentralized$_5$ & 21.7 & 14.3 & 14.2 & 15.6 & 19.5 & 17.1 & 57.9 & 48.4 & 50.2 & 52.4 & 61.7 & 54.1\\
\midrule
Local.avg. & 17.7 & 13.6 & 14.3 & 15.8 & 16.0 & 15.5 & 55.4 & 46.0 & 48.3 & 51.4 & 51.1 & 51.1 \\
\bottomrule
\end{tabular}
}
\end{table*}

\begin{table*}[ht!]
\centering
\label{tab:re-train}
\caption{The performance of the server model after 25 commu. turns and the averaged performance of corresponding local models after 25 and 26 commu. turns, on each client denoted as $\{C_i\}_{i=1}^5$. We utilize the ConVIRT as the backbone. We report the models after local update on client datasets of FedAvg pre-training strategy as Local$_i$.}
\resizebox{0.95\textwidth}{!}{
\begin{tabular}{@{} llllll llllll}
\toprule
 \multicolumn{1}{c}{\textbf{Strategy}} & \multicolumn{1}{c}{\textbf{com. turn}} & \multicolumn{5}{c}{\textbf{Recall@1 (ACC)}} & \multicolumn{5}{c}{\textbf{Recall@5 (ACC)}}\\
\midrule 
 &  & \textbf{C1} & \textbf{C2} & \textbf{C3} & \textbf{C4} & \textbf{C5} & \textbf{C1} & \textbf{C2} & \textbf{C3} & \textbf{C4} & \textbf{C5}\\ 
\midrule
FedAvg & 25 & 30.4 & 25.3 & 26.9 & 28.8 & 32.7 & 76.4 & 66.7 & 69.8 & 73.8 & 73.9\\
\midrule
Local$_0$ & 25 & 31.8 & 23.4 & 26.0 & 26.2 & 33.7 & 73.4 & 64.1 & 67.2 & 71.4 & 71.8\\
Local$_1$ & 25 & 27.7 & 22.5 & 23.8 & 25.1 & 27.7 & 73.4 & 63.1 & 64.6 & 69.2 & 68.6\\
 Local$_2$ & 25 & 28.6 & 24.4 & 24.3 & 27.3 & 28.9 & 73.3 & 64.9 & 67.5 & 70.3 & 71.8\\
 Local$_3$ & 25 & 30.6 & 22.6 & 23.9 & 25.4 & 26.4 & 72.6 & 64.0 & 66.3 & 68.9 & 67.6\\
Local$_4$ & 25 & 27.3 & 24.4 & 25.5 & 26.5 & 28.9 & 73.3 & 65.8 & 69.0 & 71.2 & 69.2\\
\midrule
1-5 Avg. & 25 & 29.2$\downarrow$ & 23.5$\downarrow$ & 24.7$\downarrow$ & 26.1$\downarrow$ & 29.1$\downarrow$ & 73.2$\downarrow$ & 64.4$\downarrow$ & 66.9$\downarrow$ & 70.2$\downarrow$ & 69.8$\downarrow$\\
\midrule
Local$_0$ & 26 & 29.9 & 23.2 & 25.2 & 26.7 & 33.1 & 73.4 & 64.1 & 67.2 & 71.4 & 71.8\\
 Local$_1$ & 26 & 28.7 & 23.1 & 24.0 & 25.9 & 27.4 & 74.5 & 63.4 & 64.1 & 69.7 & 67.9\\
Local$_2$ & 26 & 30.9 & 23.9 & 25.4 & 27.7 & 29.9 & 72.4 & 65.6 & 67.4 & 71.6 & 71.1\\
 Local$_3$ & 26 & 30.1 & 22.7 & 23.4 & 24.5 & 27.1 & 73.1 & 63.5 & 65.3 & 67.5 & 67.3\\
 Local$_4$ & 26 & 27.0 & 24.8 & 25.5 & 26.6 & 29.6 & 73.6 & 66.0 & 69.2 & 71.2 & 69.5\\
\midrule
 1-5 Avg. & 26 & 29.3$\downarrow$ & 23.5$\downarrow$ & 24.7$\downarrow$ & 26.3$\downarrow$ & 29.4$\downarrow$ & 73.3$\downarrow$ & 64.4$\downarrow$ & 66.3$\downarrow$ & 70.1$\downarrow$ & 69.4$\downarrow$\\
\bottomrule
\end{tabular}
}
\end{table*}

\begin{table*}[ht!]
\centering
\label{tab:hetero.pos.}
\caption{The accuracy of the test set of each client. We show the performance of FedAvg pre-trained baseline and its retrained models on different client datasets. We report the models after local update on client datasets of FedAvg pre-training strategy as Local$_i$.}
\resizebox{\textwidth}{!}{
\begin{tabular}{@{} llllllll lllllll}
\toprule
\textbf{position} & \multicolumn{1}{c}{\textbf{model}} & \multicolumn{1}{c}{\textbf{com.}} & \multicolumn{6}{c}{\textbf{Recall@1 (ACC)}} & \multicolumn{6}{c}{\textbf{Recall@5 (ACC)}}\\
\midrule 
 & & & \textbf{C0} & \textbf{C1} & \textbf{C2} & \textbf{C3} & \textbf{C4} & \textbf{Avg.} & \textbf{C0} & \textbf{C1} & \textbf{C2} & \textbf{C3} & \textbf{C4} & \textbf{Avg.}\\ 
\midrule
- & server & 25 & 30.4 & 25.3 & 26.9 & 28.8 & 32.7 & 28.8 & 76.4 & 66.7 & 69.8 & 73.8 & 73.9 & 72.1\\
- & server & 50 & 32.3 & 26.0 & 27.0 & 27.1 & 30.2 & 28.5$\downarrow$ & 77.6 & 67.9 & 69.4 & 72.1 & 71.7 & 71.7$\downarrow$\\
\midrule
$\rightarrow$ shallow & Local$_0$ & 25 & 30.4 & 25.0 & 25.3 & 28.4 & 28.6 & 27.5$\downarrow$ & 73.8 & 67.2 & 68.4 & 72.0 & 73.0 & 70.9$\downarrow$\\
$\rightarrow$ shallow & Local$_1$ & 25 & 34.3 & 26.4 & 27.3 & 29.7 & 30.2 & 29.4$\uparrow$ & 78.3 & 69.8 & 72.3 & 75.1 & 77.4 & 74.6$\uparrow$\\
$\rightarrow$ shallow & Local$_2$ & 25 & 33.7 & 26.4 & 27.3 & 29.7 & 30.2 & 29.4$\uparrow$ & 77.3 & 67.4 & 70.7 & 74.2 & 70.5 & 72.0$\uparrow$\\
$\rightarrow$ shallow & Local$_3$ & 25 & 27.7 & 18.9 & 19.3 & 25.6 & 24.9 & 25.0$\downarrow$ & 72.4 & 64.2 & 64.8 & 70.5 & 71.1 & 68.6$\downarrow$\\
$\rightarrow$ shallow & Local$_4$ & 25 & 26.2 & 18.9 & 19.3 & 22.7 & 22.7 & 22.0$\downarrow$ & 69.3 & 56.8 & 58.9 & 63.3 & 64.5 & 62.6$\downarrow$\\
\bottomrule
\end{tabular}
}
\end{table*}

\begin{table*}[ht!]
\centering
\label{tab:align distort}
\caption{The accuracy of the model on each client. We show the acc. of centralized and FedAvg pre-trained baselines and de-centralized pre-trained models shown as Local$_i$ retrained on the union of training splits of client datasets. We fine-tune shallow layers of the de-centralized pre-trained model with the union dataset.}
\resizebox{0.95\textwidth}{!}{
\begin{tabular}{@{} lllllll lllllll}
\toprule
\textbf{strategy} & \multicolumn{1}{c}{\textbf{model}} & \multicolumn{5}{c}{\textbf{Recall@1 (ACC)}} & \multicolumn{5}{c}{\textbf{Recall@5 (ACC)}}\\
\midrule 
 & & \textbf{C1} & \textbf{C2} & \textbf{C3} & \textbf{C4} & \textbf{C5} & \textbf{Avg.} & \textbf{C1} & \textbf{C2} & \textbf{C3} & \textbf{C4} & \textbf{C5} & \textbf{Avg.}\\ 
\midrule
Global & server & 43.6 & 38.6 & 40.1 & 43.1 & 41.9 & 41.5 & 86.6 & 80.0 & 82.6 & 85.0 & 86.8 & 84.2\\
\midrule
FedAvg & server & 30.4 & 25.3 & 26.9 & 28.8 & 32.7 & 28.8 & 76.4 & 66.7 & 69.8 & 73.8 & 73.9 & 72.1\\
\midrule
Decentralized & Local$_1$ & 28.7 & 22.6 & 23.5 & 22.3 & 24.6 & 24.4 & 70.9 & 60.8 & 63.3 & 62.4 & 63.4 & 64.2\\
Decentralized & Local$_2$ & 17.4 & 19.9 & 18.2 & 17.6 & 17.7 & 18.1 & 52.7 & 56.1 & 55.1 & 54.0 & 53.5 & 54.3\\
Decentralized & Local$_3$ & 20.9 & 20.7 & 26.0 & 21.0 & 22.3 & 22.2 & 58.9 & 58.1 & 65.3 & 58.2 & 59.0 & 59.9\\
Decentralized & Local$_4$ & 20.9 & 20.1 & 20.7 & 25.6 & 21.1 & 21.7 & 59.1 & 56.9 & 57.8 & 64.7 & 58.7 & 59.4\\
Decentralized & Local$_5$ & 21.8 & 19.5 & 22.0 & 20.9 & 31.5 & 23.2 & 60.7 & 57.0 & 59.8 & 60.2 & 74.1 & 62.4\\
\midrule
Decentralized & Local$_1$ & 17.7 & 14.7 & 15.4 & 18.2 & 14.1 & 16.0 & 57.0 & 49.6 & 51.3 & 54.6 & 55.1 & 53.5\\
Decentralized & Local$_2$ & 15.3 & 11.6 & 13.9 & 15.0 & 13.8 & 13.9 & 54.8 & 41.9 & 46.2 & 47.9 & 45.6 & 47.3\\
Decentralized & Local$_3$ & 17.4 & 13.1 & 14.1 & 15.1 & 15.4 & 15.0 & 50.4 & 44.2 & 46.4 & 49.7 & 52.5 & 48.6\\
Decentralized & Local$_4$ & 16.5 & 14.3 & 14.1 & 15.4 & 17.4 & 15.5 & 57.0 & 45.7 & 47.5 & 52.4 & 57.6 & 52.0\\
Decentralized & Local$_5$ & 21.7 & 14.3 & 14.2 & 15.6 & 19.5 & 17.1 & 57.9 & 48.4 & 50.2 & 52.4 & 61.7 & 54.1\\
\bottomrule
\end{tabular}
}
\end{table*}

\clearpage

\section{Theoretical Analysis}
\subsection{Derivation of Proposition 1}
To begin, we will establish the following lemma.
\begin{lemma}
     For a batch of samples \( z_a \) and \( z_b \) with batch size \(\text{bz}\), and temperature parameter \(\tau\). Let \(\overline{D}\) is the average L2 distance across the batch, $D_{\max}$ is the maximum L2 distance across the dataset. Suppose for all optimization batches there exist $L\geq1, 0\leq l\leq 1$, such that \( l^2 \cdot D_{\max}^2 \leq \| z_{a,i} - z_{b,i} \|_2^2 \leq L^2 \cdot D_{\max}^2 \), for $\forall z_{a,i},z_{b,i}$ in the batch. Then the contrastive loss \( L_{\text{CL}}(z_a, z_b) \) has the following upper bound:

\[
L_{\text{CL}}(z_a, z_b) \leq \log(\text{bz}) + \alpha\overline{D}^2,
\]
where $\alpha=\frac{L^2(1-l^2)}{2\tau}$, is a client-specific constant.
\end{lemma}
\begin{proof}
The contrastive loss \( L_{\text{CL}} \) is given by
\[
L_{\text{CL}}(z_a, z_b) = \frac{1}{\text{bz}} \sum_{i=1}^{\text{bz}} \left[ -\log \frac{\exp(\text{sim}(z_{a,i}, z_{b,i}) / \tau)}{\sum_{j=1}^{\text{bz}} \exp(\text{sim}(z_{a,i}, z_{b,j}) / \tau)} \right],
\]
where \( \text{sim}(z_{a,i}, z_{b,i}) = \frac{z_{a,i} \cdot z_{b,i}}{\|z_{a,i}\| \|z_{b,i}\|} \). For normalized vectors \( z_{a,i} \) and \( z_{b,i} \), we have:
\[
\| z_{a,i} - z_{b,i} \|_2^2 = 2 - 2 \cdot \text{sim}(z_{a,i}, z_{b,i}), \quad \Rightarrow \quad \text{sim}(z_{a,i}, z_{b,i}) = 1 - \frac{1}{2} \| z_{a,i} - z_{b,i} \|_2^2.
\]

Substituting \( \text{sim}(z_{a,i}, z_{b,i}) \) into \( L_{\text{CL}} \):
\[
L_{\text{CL}}(z_a, z_b) = \frac{1}{\text{bz}} \sum_{i=1}^{\text{bz}} \left[ -\log \frac{\exp\left( (1 - \frac{1}{2} \| z_{a,i} - z_{b,i} \|_2^2) / \tau \right)}{\sum_{j=1}^{\text{bz}} \exp\left( (1 - \frac{1}{2} \| z_{a,i} - z_{b,j} \|_2^2) / \tau \right)} \right].
\]

Let \( D_{\max}^2 = \max_{i} \| z_{a,i} - z_{b,i} \|_2^2 \), we have:
\[
L_{\text{CL}}(z_a, z_b) \leq \log(\text{bz}) + \frac{1}{2\tau} \left( D_{\max}^2 - \frac{1}{\text{bz}} \sum_{i=1}^{\text{bz}} \| z_{a,i} - z_{b,i} \|_2^2 \right).
\]

Substituting the assumption \( l^2 \cdot D_{\max}^2 \leq \| z_{a,i} - z_{b,i} \|_2^2 \leq L^2 \cdot D_{\max}^2 \), we get:
\[
L_{\text{CL}}(z_a, z_b) \leq \log(\text{bz}) + \frac{1}{2\tau} \left( D_{\max}^2 - \frac{1}{\text{bz}} \sum_{i=1}^{\text{bz}} l^2 \cdot D_{\max}^2 \right).
\]

This simplifies further to:
\[
L_{\text{CL}}(z_a, z_b) \leq \log(\text{bz}) + \frac{1}{2\tau} \cdot D_{\max}^2 \cdot (1 - l^2).
\]

Now, since we have \( D_{\max} \leq L \cdot \overline{D} \), where \( \overline{D} = \frac{1}{\text{bz}} \sum_{i=1}^{\text{bz}} \| z_{a,i} - z_{b,i} \|_2^2 \). Substituting this into the inequality above:
\[
L_{\text{CL}}(z_a, z_b) \leq \log(\text{bz}) + \frac{1}{2\tau} \cdot (L \cdot \overline{D})^2 \cdot (1 - l^2).
\]

Replacing $\alpha=\frac{L^2(1-l^2)}{2\tau}$, we have:
\[
L_{\text{CL}}(z_a, z_b) \leq \log(\text{bz}) + \alpha\overline{D}^2.
\]
\qed
\end{proof}

Using this lemma, we will complete the proof of Proposition 1. In this paper, without loss of generalizability, we assume $\mathcal{R}_{T}(\cdot)$ to be the contrastive loss.

\begin{proof}
We begin by expressing the generalization error \( \mathcal{R}_T(\hat{f}) \) on the target domain \( \mathcal{D}_{\mathcal{T}} \) as the expected contrastive loss:
\begin{equation} \label{eq:generalization_error}
\mathcal{R}_T(\hat{f}) = \mathbb{E}_{(x, y) \sim \mathcal{D}_{\mathcal{T}}} \left[ L_{\text{CL}}( \hat{f}_{\psi}(x), \hat{f}_{\phi}(y) ) \right],
\end{equation}
where \( L_{\text{CL}} \) is the contrastive loss defined as:
\begin{equation} \label{eq:contrastive_loss}
L_{\text{CL}}(z_a, z_b) = \frac{1}{\text{bz}} \sum_{i=1}^{\text{bz}} \left[ -\log \frac{\exp\left( \text{sim}(z_{a,i}, z_{b,i}) / \tau \right)}{\sum_{j=1}^{\text{bz}} \exp\left( \text{sim}(z_{a,i}, z_{b,j}) / \tau \right)} \right],
\end{equation}
with \( \text{sim}(z_{a,i}, z_{b,i}) = \frac{z_{a,i}^\top z_{b,i}}{\| z_{a,i} \| \| z_{b,i} \|} \) and \( \text{bz} \) being the batch size.

By Lemma 1, we have an upper bound on the contrastive loss:
\begin{equation} \label{eq:lemma_bound}
L_{\text{CL}}(z_a, z_b) \leq \log(\text{bz}) + \alpha\overline{D}^2,
\end{equation}
where \( \overline{D}^2 \) is the average squared Euclidean distance between \( z_{a,i} \) and \( z_{b,i} \):
\begin{equation} \label{eq:average_distance}
\overline{D}^2 = \frac{1}{\text{bz}} \sum_{i=1}^{\text{bz}} \| z_{a,i} - z_{b,i} \|_2^2.
\end{equation}

Applying this to our generalization error:
\begin{equation} \label{eq:generalization_error_bound}
\mathcal{R}_T(\hat{f}) \leq \log(\text{bz}) + \alpha \mathbb{E}_{(x, y) \sim \mathcal{D}_{\mathcal{T}}} \left[ \| \hat{f}_{\psi}(x) - \hat{f}_{\phi}(y) \|_2^2 \right].
\end{equation}

Then, we have:
\begin{align}
\label{eq_main}
| \hat{f}_{\psi}(x) - \hat{f}_{\phi}(y) |_2^2 &\leq \left( | \hat{f}_{\psi}(x) - \hat{f}_{\psi_i}(x) |_2 + | \hat{f}_{\psi_i}(x) - f_{\psi_i}(x) |_2 \right. \nonumber \\
&\quad \left. + | f_{\psi_i}(x) - f_{\phi_i}(y) |_2 + | f_{\phi_i}(y) - \hat{f}_{\phi_i}(y) |_2 + | \hat{f}_{\phi_i}(y) - \hat{f}_{\phi}(y) |_2 \right)^2 \nonumber \\
&\leq 5 \left( | \hat{f}_{\psi}(x) - \hat{f}_{\psi_i}(x) |_2^2 + | \hat{f}_{\psi_i}(x) - f_{\psi_i}(x) |_2^2 \right. \nonumber \\
&\quad \left. + | f_{\psi_i}(x) - f_{\phi_i}(y) |_2^2 + | f_{\phi_i}(y) - \hat{f}_{\phi_i}(y) |_2^2 + | \hat{f}_{\phi_i}(y) - \hat{f}_{\phi}(y) |_2^2 \right),
\end{align}
where the last inequality follows from the fact that for any real numbers $a_1, \dots, a_n$,
\[
\left( \sum_{j=1}^n a_j \right)^2 \leq n \sum_{j=1}^n a_j^2.
\]
Define the error terms:
\begin{align*}
\epsilon_{\hat{f}_{\psi}, \hat{f}_{\psi_i}}(x) &= | \hat{f}_{\psi}(x) - \hat{f}_{\psi_i}(x) |_2^2, \\
\epsilon_{\hat{f}_{\psi_i}, f_{\psi_i}}(x) &= | \hat{f}_{\psi_i}(x) - f_{\psi_i}(x) |_2^2, \\
\epsilon_{\hat{f}_{\phi}, \hat{f}_{\phi_i}}(y) &= | \hat{f}_{\phi}(y) - \hat{f}_{\phi_i}(y) |_2^2, \\
\epsilon_{\hat{f}_{\phi_i}, f_{\phi_i}}(y) &= | \hat{f}_{\phi_i}(y) - f_{\phi_i}(y) |_2^2, \\
C_i(x, y) &= | f_{\psi_i}(x) - f_{\phi_i}(y) |_2^2.
\end{align*}

Then inequality \eqref{eq_main} becomes:
\begin{equation}
\label{eq_modified}
| \hat{f}_{\psi}(x) - \hat{f}_{\phi}(y) |_2^2 \leq 5 \left( \epsilon_{\hat{f}_{\psi}, \hat{f}_{\psi_i}}(x) + \epsilon_{\hat{f}_{\psi_i}, f_{\psi_i}}(x) + C_i(x, y) + \epsilon_{\hat{f}_{\phi_i}, f_{\phi_i}}(y) + \epsilon_{\hat{f}_{\phi}, \hat{f}_{\phi_i}}(y) \right).
\end{equation}

Taking expectation over $(x, y) \sim \mathcal{D}_{\mathcal{T}}$ and using $\mathcal{D}_{\mathcal{T}} = \sum_{i=1}^N w_i \mathcal{D}_i$, we have:
\begin{align}
\mathbb{E}_{(x, y) \sim \mathcal{D}_{\mathcal{T}}} \left[ | \hat{f}_{\psi}(x) - \hat{f}_{\phi}(y) |_2^2 \right] &= \sum_{i=1}^N w_i \, \mathbb{E}_{(x, y) \sim \mathcal{D}_i} \left[ | \hat{f}_{\psi}(x) - \hat{f}_{\phi}(y) |_2^2 \right] \nonumber \\
&\leq 5 \sum_{i=1}^N w_i \, \mathbb{E}_{(x, y) \sim \mathcal{D}_i} \left[ \epsilon_{\hat{f}_{\psi}, \hat{f}_{\psi_i}}(x) + \epsilon_{\hat{f}_{\psi_i}, f_{\psi_i}}(x) \right. \nonumber \\
&\quad \left. + C_i(x, y) + \epsilon_{\hat{f}_{\phi_i}, f_{\phi_i}}(y) + \epsilon_{\hat{f}_{\phi}, \hat{f}_{\phi_i}}(y) \right].
\end{align}

Define $C_i = \mathbb{E}_{(x, y) \sim \mathcal{D}_i} \left[ C_i(x, y) \right]$. Then,
\begin{align}
\label{eq_expectation}
\mathbb{E}_{(x, y) \sim \mathcal{D}_{\mathcal{T}}} \left[ | \hat{f}_{\psi}(x) - \hat{f}_{\phi}(y) |_2^2 \right] &\leq 5 \sum_{i=1}^N w_i \left( \mathbb{E}_{x \sim \mathcal{D}_i} \left[ \epsilon_{\hat{f}_{\psi}, \hat{f}_{\psi_i}}(x) + \epsilon_{\hat{f}_{\psi_i}, f_{\psi_i}}(x) \right] \right. \nonumber \\
&\quad \left. + \mathbb{E}_{y \sim \mathcal{D}_i} \left[ \epsilon_{\hat{f}_{\phi_i}, f_{\phi_i}}(y) + \epsilon_{\hat{f}_{\phi}, \hat{f}_{\phi_i}}(y) \right] + C_i \right).
\end{align}

Substituting \eqref{eq_expectation} into the generalization error bound, we obtain:
\begin{align}
\mathcal{R}_T(\hat{f}) &\leq \log(\text{bz}) + \alpha \sum_{i=1}^N w_i \left( 5 \mathbb{E}_{x \sim \mathcal{D}_i} \left[ \epsilon_{\hat{f}_{\psi}, \hat{f}_{\psi_i}}(x) + \epsilon_{\hat{f}_{\psi_i}, f_{\psi_i}}(x) \right] \right. \nonumber \\
&\quad \left. + 5 \mathbb{E}_{y \sim \mathcal{D}_i} \left[ \epsilon_{\hat{f}_{\phi_i}, f_{\phi_i}}(y) + \epsilon_{\hat{f}_{\phi}, \hat{f}_{\phi_i}}(y) \right] + 5 C_i \right).
\end{align}

Letting $\alpha_i = 5 \alpha$, we have:
\begin{equation}
\mathcal{R}_T(\hat{f}) \leq \sum_{i=1}^N w_i \alpha_i \left( \mathbb{E}_{x \sim \mathcal{D}_i} \left[ \epsilon_{\hat{f}_{\psi}, \hat{f}_{\psi_i}}(x) + \epsilon_{\hat{f}_{\psi_i}, f_{\psi_i}}(x) \right] + \mathbb{E}_{y \sim \mathcal{D}_i} \left[ \epsilon_{\hat{f}_{\phi_i}, f_{\phi_i}}(y) + \epsilon_{\hat{f}_{\phi}, \hat{f}_{\phi_i}}(y) \right] + C_i \right) + \log(\text{bz}).
\end{equation}

Since $\log(\text{bz})$ is independent of $\hat{f}$, it can be considered a constant. Thus, we can express the generalization error as:
\[
\mathcal{R}_T(\hat{f}) \leq \sum_{i=1}^N w_i \alpha_i \left( \epsilon_{\hat{f}_{\psi}, \hat{f}_{\psi_i}} + \epsilon_{\hat{f}_{\psi_i}, f_{\psi_i}} + \epsilon_{\hat{f}_{\phi}, \hat{f}_{\phi_i}} + \epsilon_{\hat{f}_{\phi_i}, f_{\phi_i}} + C_i \right).
\]
This completes the proof of Proposition 1.
\end{proof}

\end{document}